\documentclass[10pt,twocolumn,letterpaper]{article}

\usepackage[pagenumbers]{cvpr} 

\newcommand{\nsubjects}{$29$\xspace}
\newcommand{\ndatasets}{$6$\xspace}

\newcommand{\oursnamefull}{High-Dimensional Gaussian Splatting \\for High-Fidelity Animatable Face Avatars}
\newcommand{\oursname}{HyperGaussian\xspace}
\newcommand{\ourslongname}{HyperGaussian\xspace}

\newcommand{\R}{\mathbb{R}}
\newcommand{\bs}{\boldsymbol}
\newcommand{\NN}{\mathcal{N}}
\DeclareMathOperator{\tr}{tr}
\DeclareMathOperator*{\argmax}{\arg\max}

\definecolor{psnrcolor}{RGB}{161,102,170}
\definecolor{ssimcolor}{RGB}{251,176,64}
\definecolor{lpipscolor}{RGB}{102,191,115}

\usepackage{adjustbox}
\usepackage{pgffor}
\usepackage{float}
\usepackage[table]{xcolor}

\definecolor{cvprblue}{rgb}{0.21,0.49,0.74}
\usepackage[pagebackref,breaklinks,colorlinks,allcolors=cvprblue]{hyperref}

\def\paperID{*****} 
\def\confName{CVPR}
\def\confYear{2026}

\title{\oursname{}s: \oursnamefull}

\author{Gent Serifi \quad Marcel C. Buehler\vspace{.5em}\\
ETH Zurich, Switzerland\\
\small{\url{https://gserifi.github.io/HyperGaussians}}
}

\begin{document}

\twocolumn[{
\maketitle
\begin{center}
    \centering
    \captionsetup{type=figure}
    \includegraphics[width=1\textwidth]{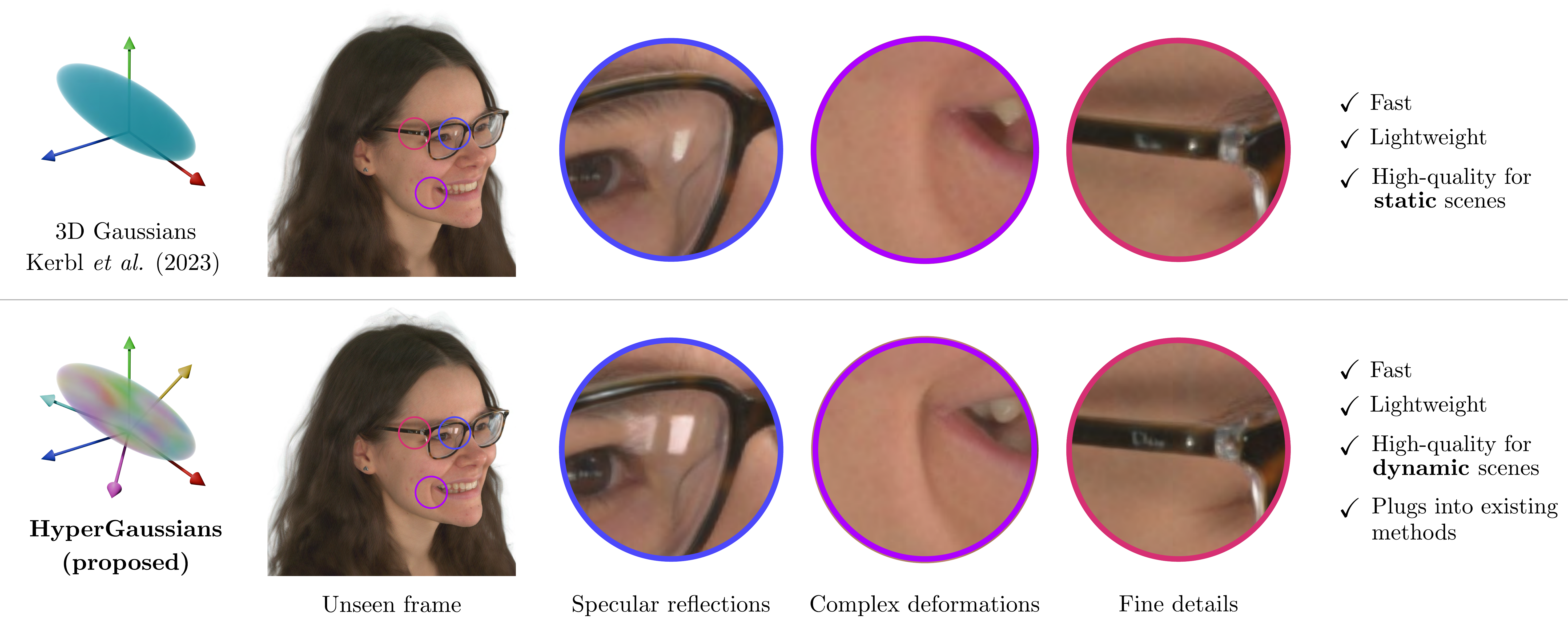}
    \caption{\label{fig:teaser}We propose a plug-and-play enhancement for 3D Gaussians:  \emph{\oursname{}s}. \oursname{}s extend 3D Gaussian Splatting to higher dimensions, resulting in improved high-frequency details for specular reflections and thin structures. \oursname{}s can be easily integrated into existing models with minimal overhead. This figure shows the effect of plugging \oursname{}s (bottom) into the state-of-the-art for multi-view face avatars, GaussianHeadAvatar \cite{xu2024gaussianheadavatar} (top). Note the improvement in specular reflections, complex deformations, and fine details. \cref{fig:qualitative_comparison} shows results for monocular face avatars \cite{xiang2024flashavatar}.}
\end{center}
}]
\begin{abstract}
\noindent
We introduce HyperGaussians, a novel extension of 3D Gaussian Splatting for high-quality animatable face avatars. 
While tremendous successes have been achieved for static faces, animatable avatars from dynamic videos still fall in the uncanny valley. The de facto standard, 3D Gaussian Splatting (3DGS), represents a face through a collection of 3D Gaussian primitives. 3DGS excels at rendering static faces, but the state-of-the-art still struggles with nonlinear deformations, complex lighting effects, and fine details. While most related works focus on predicting better Gaussian parameters from expression codes, we rethink the 3D Gaussian representation itself and how to make it more expressive. Our insights lead to a novel extension of 3D Gaussians to high-dimensional multivariate Gaussians, dubbed 'HyperGaussians'. The higher dimensionality increases expressivity through conditioning on a learnable local embedding. However, splatting HyperGaussians is computationally expensive because it requires inverting a high-dimensional covariance matrix. We solve this by reparameterizing the covariance matrix, dubbed the 'inverse covariance trick'. This trick boosts the efficiency so that HyperGaussians can be seamlessly integrated into existing models. To demonstrate this, we plug in HyperGaussians into two state-of-the-art methods for face avatars: FlashAvatar and GaussianHeadAvatar. Our evaluation on 29 subjects from 6 face datasets shows that HyperGaussians outperform 3DGS numerically and visually, particularly for high-frequency details like eyes, teeth, wrinkles, and specular reflections.
\end{abstract}
\begin{figure*}[t]
  \centering
  \includegraphics[width=1.0\linewidth]{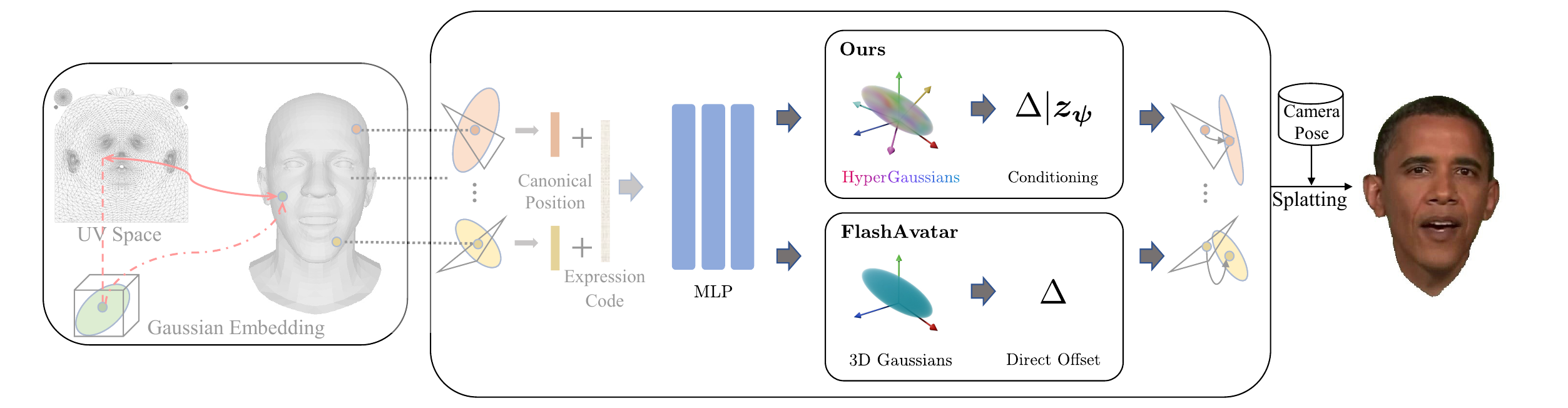}
  \caption{We propose an expressive extension to 3D Gaussians, dubbed \oursname{}s, and plug them into existing methods for face avatars, FlashAvatar \cite{xiang2024flashavatar} and GaussianHeadAvatar \cite{xu2024gaussianheadavatar}. FlashAvatar modulates 3D Gaussian primitives with expression-dependent offsets $\Delta$. We make a single modification to the pipeline: plugging \ourslongname{}s (\cref{ssec:multivariate_gaussians}) in between the MLP output and the rasterization, which modifies the offsets $\Delta$ in higher dimensions.  Instead of directly predicting offsets $\Delta$, we predict a latent $z_\psi$ that conditions \ourslongname{}s. Without \emph{any other modifications} or hyperparameter tuning, this simple change leads to a performance boost in rendering high-frequency details in the final avatar (\cref{tab:quantitative_comparison} and \cref{fig:qualitative_comparison}). \emph{This figure has been adapted from FlashAvatar \cite{xiang2024flashavatar}.}\vspace{-1em}
  \label{fig:method_overview}}
\end{figure*}

\section{Introduction}
\label{sec:intro}
Modeling dynamic scenes, such as human faces, is a long-standing problem in 3D Computer Vision and has a variety of applications in augmented and virtual reality, the entertainment industry, and virtual telepresence \cite{latoschik2017effect,orts2016holoportation}.
Faces exhibit highly nonlinear deformations such as eye blinks and mouth opening, and complex lighting effects like specular reflections in the eyes and on glasses \cite{li2022eyenerf,li2024shellnerf,saito2024relightable,sarkar2023litnerf}. Faithfully modeling these details is essential to overcome the uncanny valley, enhancing human-avatar interaction, and creating truly lifelike avatars \cite{mori1970bukimi,katsyri2015review}. One crucial element towards this goal is the development of high-quality and efficient 3D representations.

Recently, 3D Gaussian Splatting \cite{kerbl2023_3dgs} has emerged as the standard for modeling humans due to its quality and efficiency \cite{shao2024splattingavatar,chen2024monogaussianavatar,xiang2024flashavatar,zielonka2025synthetic,saito2024relightable,xu2024gaussianheadavatar}. The state-of-the-art for human faces, rigs Gaussians with Morphable Face Models \cite{blanz1999morphable,gerig2018morphablebfm2017,FLAME:SiggraphAsia2017}. Each Gaussian is attached to a mesh triangle and follows its deformation, for example, using linear blend skinning \cite{FLAME:SiggraphAsia2017,SMPL:2015,lewis2023pose}. Such linear models are a good approximation, but they cannot represent nonlinear deformations and specular effects. To mitigate this, recent work leverages neural networks to predict offsets from an input expression \cite{xiang2024flashavatar,shao2024splattingavatar,chen2024monogaussianavatar,xu2024gaussianheadavatar}, improving results over static Gaussians. Yet, \cref{fig:qualitative_comparison} and \cref{fig:gha-qualitative} show that these methods struggle to represent thin structures like teeth and hair, specular reflections on eyes and glasses, and nonlinear deformations such as closing eyelids. These high-frequency details are key to achieving believable digital human avatars.

In this work, we extend 3D Gaussian Splatting \cite{kerbl2023_3dgs} to arbitrary higher dimensions and call the novel representation \emph{\ourslongname{}s} (\cref{ssec:multivariate_gaussians}). \ourslongname{}s are an expressive, high-dimensional representation for high-quality face avatars. We motivate \ourslongname{}s by analyzing them through a probabilistic lens (\cref{ssec:multivariate_gaussians}). At their core, they are multivariate Gaussians that parameterize a conditional (with respect to local embeddings) distribution over vanilla 3DGS scenes. The additional dimensions and local embeddings improve complex local deformations and specular effects. However, due to the higher dimensionality, a na\"ive implementation of high-dimensional Gaussians requires substantial computational resources for splatting (\cref{fig:conditioning_benchmark}). To solve this, we propose a re-parameterization, called the \emph{inverse covariance trick} (\cref{ssec:inverse_covariance_trick}). Using the inverse covariance trick, high-dimensional Gaussian primitives can be splatted efficiently in real time at 300 FPS. 

To the best of our knowledge, we are the first to improve high-frequency details in facial avatars by enhancing the low-level parameterization of the Gaussian primitives. Most related works focus on improving a model architecture or training scheme \cite{xiang2024flashavatar,chen2024monogaussianavatar,shao2024splattingavatar,qian2024gaussianavatars,lee2025surfhead,xu2024gaussianheadavatar} that estimates the dynamics of the Gaussian parameters using neural networks. Our contribution, \ourslongname{}s, is orthogonal to these efforts. \ourslongname{}s build on top of the 3DGS framework and thus can be readily plugged into an existing pipeline and boost its performance out-of-the-box, \emph{without any other changes} to the model, architecture, or hyperparameters. We demonstrate this by integrating our proposed \ourslongname{}s into the state-of-the-art in fast face avatar learning from monocular video, FlashAvatar \cite{xiang2024flashavatar}, and the state-of-the-art for multi-view video, GaussianHeadAvatar \cite{xu2024gaussianheadavatar}.
The only change required is to replace the 3D Gaussians with \ourslongname{}s, the rest stays \emph{exactly} the same. We evaluate on videos of \nsubjects subjects from \ndatasets different datasets. The result is a boost in performance for self- and cross-reenactment, as demonstrated in \cref{fig:teaser,fig:qualitative_comparison,fig:cross_reenactment,fig:gha-qualitative} and \cref{tab:quantitative_comparison}. Specifically, \ourslongname{}s significantly improve thin structures like glass frames and teeth, specular reflections on eyes and glasses, and complex nonlinear deformations such as closing of the eyelid and wrinkles.

In summary, this paper contributes the following:
\begin{itemize}
    \item \emph{\oursname{}s}, a novel, expressive representation for modeling high-frequency and dynamic effects supported by a probabilistic interpretation.
    \item The \emph{inverse covariance trick}, a technical contribution that yields a substantial improvement in computational efficiency of high-dimensional Gaussian conditioning.
    \item Evaluation of \oursname{}s applied to the state-of-the-art for fast face avatars, FlashAvatar, and multi-view avatars, GaussianHeadAvatar, demonstrating improved rendering quality without \emph{any other changes} across \nsubjects subjects from \ndatasets datasets.
\end{itemize}

\section{Related Work}
\label{sec:related_work}

\paragraph{Dynamic 3D Scenes}
Neural 3D representations like Neural Radiance Fields (NeRFs) \cite{mildenhall2020nerf}, Neural Implicit Functions \cite{park2019deepsdf,mescheder2019occupancy}, and 3D Gaussians \cite{kerbl2023_3dgs} can represent 3D scenes with a high level of detail and enable photorealistic novel view synthesis. 
NeRFs and Implicit Functions can be conditioned to model dynamic effects over time \cite{park2021nerfies,park2021hypernerf,pumarola2021d} or facial expressions \cite{gafni2021dynamic}. NeRFies \cite{park2021nerfies} optimizes a continuous deformation field that maps from a deformed to a canonical space, but it suffers from artifacts when topological changes occur, \eg, a mouth opening. HyperNeRF \cite{park2021hypernerf} shows that these issues can be mitigated by modeling the deformation field in higher dimensions and \emph{slicing} through it using learnable embeddings. Inspired by HyperNeRF, we propose an extension to 3D Gaussian Splatting by modeling Gaussians in a higher-dimensional space.

\paragraph{Gaussian Splatting} 
3D Gaussian Splatting (3DGS) \cite{kerbl2023_3dgs} has effectively become the standard for modeling static scenes. Several works have explored extensions to non-standard dimensionalities \cite{huang20242d,wu20244d,diolatzis2024n,guedon2023sugar,zhou2023groomgen,luo2024gaussianhair,zakharov2024human}.
1D Cylindrical Gaussian Splatting is being used for hair modeling 
\cite{zhou2023groomgen,luo2024gaussianhair,zakharov2024human}. 2DGS \cite{huang20242d} uses oriented planar Gaussian disks to improve geometric consistency in Gaussian Splatting.
 4DGS \cite{wu20244d} extends 3DGS for rendering dynamic scenes by proposing a custom CUDA kernel for splatting 4D Gaussians with a time dimension. NDGS \cite{diolatzis2024n} formulates a Gaussian Mixture Model for representing static scenes with high appearance variability, which is effective for highly reflective surfaces. Particular instances of NDGS, such as 6DGS~\cite{gao20246dgs} and 7DGS~\cite{gao20257dgs}, use the additional dimensions to augment the Gaussians with view direction and time. We propose a novel, unified probabilistic interpretation of these efforts and identify critical limitations in their representation (\cref{ssec:multivariate_gaussians}). We address these shortcomings and highlight two key differences between their representations and \ourslongname{}s. First, their representation has no degrees of freedom for the conditional orientation of the 3D Gaussians, \emph{i.e.}, the Gaussians cannot rotate. Second, the rendered size of a Gaussian depends on the probability density function of the joint distribution. This causes Gaussians to disappear under large deformations, leading to semantic subparts of the scene being modeled by multiple Gaussians that are invisible most of the time. Our formulation removes these limitations and ensures that the Gaussians deform consistently. Furthermore, NDGS-based methods are limited by the computational effort required to condition their primitives. This makes them unsuitable for high-dimensional applications. Our \textit{inverse covariance trick} overcomes this issue, as shown in \cref{fig:conditioning_benchmark}, and thus provides a more scalable framework. We compare with NDGS in the supplementary material.

\paragraph{Face Avatars} A popular goal is reconstructing a face avatar \cite{li2024uravatar,cao2022authentic,ma2024gaussianblendshapes,dhamo2024headgas,li2024talkinggaussian,varitex,buehler2024cafca,buhler2023preface,zheng2022imavatar,kirschstein2025avat3r,shivangi2025scaffoldavatar} with high fidelity and rendering it under a novel pose and expression. Most recent methods \cite{xiang2024flashavatar,xu2024gaussianheadavatar,li2024uravatar,saito2024relightable,ma2024gaussianblendshapes,teotia2024gaussianheads,xu2024gphm,song2024tri,zheng2023pointavatar,shivangi2025scaffoldavatar,li2025rgbavatar,feng2025gpavatar,wang20253d} render avatars with Gaussian Splatting. The state-of-the-art \cite{xiang2024flashavatar,shao2024splattingavatar,lee2025surfhead,qian2024gaussianavatars,xu2024gaussianheadavatar,shivangi2025scaffoldavatar,li2025rgbavatar,feng2025gpavatar,wang20253d} attaches Gaussians to the mesh of a Morphable Model (3DMM) \cite{gerig2018morphablebfm2017,blanz1999morphable,FLAME:SiggraphAsia2017}. 3DMMs serve as strong shape priors and have already been extensively used in implicit avatars, such as NerFACE~\cite{gafni2021dynamic} and INSTA~\cite{zielonka2022insta}. This enables driving the avatar with controlled expressions and head poses through linear blend skinning \cite{lewis2023pose,FLAME:SiggraphAsia2017}. SplattingAvatar \cite{shao2024splattingavatar} improves over vanilla blend skinning by optimizing embeddings on the mesh. However, SplattingAvatar does not account for the fact that deformations also affect appearance. For example, a change in head pose will displace the specular reflections on glasses (\cref{fig:qualitative_comparison}), leading to a blurry rendering.
This can be improved by predicting expression-dependent Gaussian parameter offsets \cite{xiang2024flashavatar,zielonka2025synthetic,dhamo2024headgas,chen2024monogaussianavatar,xu2024gaussianheadavatar,shivangi2025scaffoldavatar}, allowing the Gaussian parameters to change for different expressions.

In highly sparse settings, recent work proposes large, pre-trained priors for predicting Gaussian attributes in a feed-forward manner or through pseudo ground-truth fitting using multi-view generative models \cite{he2025lam,chu2024generalizable,liang2025fastavatar,tang2025gaf,kirschstein2025avat3r,taubner2025cap4d,buhler2025dream}. However, the resource-intensive pre-training renders adaptations of these methods infeasible in low-compute environments. We therefore focus on optimization-based techniques in our analysis.

In this paper, we boost high-frequency details in state-of-the-art face avatars \cite{xiang2024flashavatar,xu2024gaussianheadavatar} by replacing the vanilla 3D Gaussians \cite{kerbl2023_3dgs} with a higher-dimensional representation.
\section{Method}
\label{sec:method}
We propose a novel representation for modeling dynamic 3D scenes and apply it to face avatars. Our novel representation extends 3D Gaussians \cite{kerbl2023_3dgs} to \emph{higher dimensions} and augments each Gaussian primitive with the ability to adapt according to local embeddings. This enables the representation of finer details, such as specular reflections on glasses, skin wrinkles, and text labels (\cref{fig:teaser}). We dub our novel representation \emph{\ourslongname{}s}.

In \cref{ssec:preliminaries}, we reiterate on 3DGS before introducing \oursname{}s in \cref{ssec:multivariate_gaussians}, where we additionally present the Bayesian view, as well as the \emph{inverse covariance trick} to massively reduce compute and memory load, making \oursname{}s highly scalable. \cref{ssec:casestudy} shows how we integrate \oursname{}s into the existing methods FlashAvatar \cite{xiang2024flashavatar} and GaussianHeadAvatar \cite{xu2024gaussianheadavatar} to boost their performance (\cref{tab:quantitative_comparison}) with \emph{minimal} modifications.

\subsection{Preliminary: 3D Gaussian Splatting}
\label{ssec:preliminaries}
\label{ssec:3d_gaussian_splatting}
3D Gaussian Splatting (3DGS)~\cite{kerbl2023_3dgs} models a static scene with colored anisotropic 3D Gaussians. The Gaussians are parameterized by their mean ${\bs\mu_{\mathrm{3D}} \in \R^3}$ and covariance matrix ${\bs\Sigma_{\mathrm{3D}} \in \R^{3\times 3}}$ and their appearance is represented by their opacity $\alpha$ and a color $\bs c$, where common choices are \texttt{RGB} color or Spherical Harmonics for view-dependent effects. Given images, camera parameters, and a sparse point cloud estimate, 3DGS can reconstruct a scene by optimizing a set of Gaussians via differentiable rasterization.

To ensure that the covariance matrices remain positive semi-definite during optimization, 3DGS first defines a parametric ellipsoid using a scaling matrix $\bs S$ and a rotation matrix $\bs R$, then constructs the covariance matrix as
\begin{equation}
\label{eq:3dgs_covmat_factorization}
    \bs\Sigma_{\mathrm{3D}} = \bs R\bs S\bs S^\top\bs R^\top.
\end{equation}
These matrices are themselves parameterized by a scaling vector ${\bs s \in \R^{3}}$ and a unit quaternion ${\bs q \in \R^4}$, respectively. We denote the set of Gaussians as $\{(\bs\mu_{\mathrm{3D}}^i, \bs s^i, \bs q^i, \bs c^i, \alpha^i)\}_{i=1}^{n}$.

Novel views can be rendered by \textit{splatting} the Gaussians, the Gaussians are first projected onto the camera plane and then alpha blended with respect to their evaluated density $G(\bs x) = \exp\bigl(-\frac{1}{2}(\bs x-\bs\mu_{\mathrm{3D}})^\top\bs\Sigma^{-1}(\bs x-\bs\mu_{\mathrm{3D}})\bigr)$ and learned opacity $\alpha$.

\subsection{\ourslongname{}s}
\label{ssec:multivariate_gaussians}

\noindent\oursname{}s are an extension of 3D Gaussians to higher dimensions. As an example, a vanilla Gaussian in Kerbl \etal \cite{kerbl2023_3dgs} has $m=3$ dimensions for the mean $\bs \mu_{\mathrm{3D}}$ representing its position. Intuitively, our \ourslongname{} generalizes the vanilla Gaussian primitive to $(m + n)$ dimensions. We call $m$ the \emph{attribute} dimensionality and the additional dimensions $n$ the \emph{latent} dimensionality. In the following, we use subscripts $a$ and $b$ to indicate general random quantities. In practice, we identify them with concrete 3D Gaussian attributes and local latent embeddings, respectively. We later show that high-dimensional extensions of 3DGS admit a probabilistic interpretation.

\paragraph{Formulation} This paragraph formally describes \ourslongname{}s. Consider a general random vector ${\bs\gamma\sim\NN \bigl(\bs\mu, \bs\Sigma\bigr)}$ 
that is partitioned into vectors ${\bs\gamma = (\bs\gamma_a, \bs\gamma_b)^\top}$,
where $\bs\gamma_a \in\R^m$ 
and $\bs\gamma_b \in\R^n$.

The partitioning of $\bs\gamma$ leads to the following block matrix view of $\bs\mu$ and $\bs\Sigma$:
\begin{equation}
\label{eq:block_matrix_regular_cov}
    \bs\mu =
        \begin{bmatrix}
        \bs\mu_{a} \\ \bs\mu_{b}
        \end{bmatrix}
    ,\quad
    \bs\Sigma =
        \begin{bmatrix}
            \bs\Sigma_{aa} & \bs\Sigma_{ab} \\
            \bs\Sigma_{ba} & \bs\Sigma_{bb}
        \end{bmatrix},
\end{equation}
with $\bs\Sigma_{ba} = \bs\Sigma_{ab}^\top$. In our model, $\bs\gamma_a$ is instantiated with 3D Gaussian attributes $\mathcal A$ (\eg, position $\mathcal A_{\bs\mu_{\mathrm{3D}}}$) and $\bs\gamma_b$ with a latent code $\bs z$.
Following NDGS~\cite{diolatzis2024n}, we parameterize the mean $\bs\mu$ directly and decompose the covariance $\bs\Sigma$ into its Cholesky factor $\bs L$, such that $\bs\Sigma = \bs L\bs L^\top$, where $\bs L$ is a lower triangular matrix with positive diagonal entries. Training requires optimizing the parameters of the Cholesky factor $\bs L$ together with the mean $\bs \mu$. Please read on to the end of this section for a detailed explanation of which parameters require training.

\paragraph{Splatting} 
\ourslongname{}s can be splatted with the differentiable rasterizer proposed by Kerbl \etal \cite{kerbl2023_3dgs} after being reduced to the attribute dimensionality. This can be done by \emph{conditioning} on the latent dimensions. Geometrically, conditioning corresponds to taking an $m$-dimensional slice through the multivariate Gaussian with $(m+n)$ total dimensions. More formally, we compute the conditional distribution ${p(\bs\gamma_a\vert\bs\gamma_b) = \NN\bigl(\bs\mu_{a\vert b}, \bs\Sigma_{a\vert b}\bigr)}$. Since \ourslongname{}s follow a multivariate Gaussian distribution, the  conditional distribution can be computed in closed form \cite{bishop2006pattern}:
\begin{equation}
\begin{split}
\label{eq:conditioning_regular_cov}
    \bs\mu_{a\vert b} &=
        \bs\mu_{a} +
        \bs\Sigma_{ab}
        \bs\Sigma_{bb}^{-1}
        (\bs\gamma_b - \bs\mu_{b})
    \\
    \bs\Sigma_{a\vert b} &=
        \bs\Sigma_{aa} -
        \bs\Sigma_{ab}
        \bs\Sigma_{bb}^{-1}
        \bs\Sigma_{ba}.
\end{split}
\end{equation}

We use the mean of this conditional distribution to recover a vanilla 3D Gaussian, which can be splatted with the differentiable rasterizer introduced by Kerbl \etal \cite{kerbl2023_3dgs}. Note that for \oursname{}s, the conditional covariance matrix is unused during splatting, contrary to NDGS \cite{diolatzis2024n}, and instead offers a way to derive probabilistic quantities, such as uncertainty (see the supp. mat.).

\paragraph{Bayesian Interpretation}
A key differentiation between HyperGaussians and existing multi-dimensional Gaussians \cite{diolatzis2024n,gao20246dgs,gao20257dgs} is that our formulation is strictly more expressive. To show this, we interpret high-dimensional variants of 3DGS as probabilistic descriptors of their underlying 3D primitives. More formally, they model 3D Gaussian attributes $\mathcal A$ as random variables that are coupled to a random latent vector $\bs z$ via a joint multivariate normal distribution $p(\mathcal A, \bs z)$. This relationship is learned during training. Due to the properties of multivariate Gaussians, the coupling induces a posterior density $p(\mathcal A\vert\bs z)$ which is again Gaussian. Computing the conditional mean (\emph{i.e.}, the reduction to 3D primitives for splatting) is akin to maximum a posteriori (MAP) estimation, since
\begin{equation}
    \mathbb{E}[\mathcal A\vert\bs z] = \argmax_{\mathcal A} p(\mathcal A\vert\bs z)
\end{equation}
for Gaussian distributions. Most importantly, the associated Gaussian prior $p(\mathcal A)$ acts as an implicit regularizer, which assists in preserving details during test-time inference.

We therefore identify NDGS-based methods as well-behaved distributions over 3DGS scenes. To the best of our knowledge, we are the first to highlight this connection and provide theoretical justification for high-dimensional Gaussian Splatting. This view further allows us to analyze current limitations of NDGS~\cite{diolatzis2024n}, which \ourslongname{}s aim to mitigate. In particular, NDGS uses ${\bs\mu_{\mathrm{3D}} \gets \mathbb{E}[\mathcal{A}_{\bs\mu_{\mathrm{3D}}}\vert\bs z]}$, ${\bs\Sigma_{\mathrm{3D}}\gets\mathrm{Cov}[\mathcal{A}_{\bs\mu_{\mathrm{3D}}}\vert\bs z]}$, and $p(\mathcal{A}_{\bs\mu_{\mathrm{3D}}}, \bs z)$ to modulate the opacity. In other words, NGDS only models the dependence between the position and the latent code, while the scaling and rotation are independent because the conditional covariance matrix does not consider the realization of $\bs z$ (see \cref{eq:conditioning_regular_cov}, $\bs z$ corresponds to $\gamma_{b}$). On the other hand, \ourslongname{}s additionally model $p(\mathcal A_{\bs s}\vert\bs z)$ and $p(\mathcal A_{\bs r}\vert\bs z)$, allowing the 3D primitives to scale and rotate according to the latent code. This means that our parameterization accounts for conditional distributions of 3DGS scenes with strictly larger support than NDGS.

\paragraph{Inverse Covariance Trick for Fast Conditioning}
\label{ssec:inverse_covariance_trick}

A na\"ive implementation of the conditioning in \cref{eq:conditioning_regular_cov}, as in NDGS~\cite{diolatzis2024n}, is very inefficient for large latent codes $\bs\gamma_b$. The bottleneck lies in storing and inverting the conditional covariance matrix $\bs\Sigma_{bb}\in\R^{n\times n}$. The latent dimensionality $n$ is typically much larger than the dimensionality after conditioning $m$ ($n \gg m$).

We propose the \emph{inverse covariance trick} that substantially reduces the cost of conditioning from $\mathcal O\bigl(n^3 + mn^2\bigr)$ to $\mathcal O\bigl(m^3 + m^2n\bigr)$, a reduction from \emph{cubic} to \emph{linear} growth with respect to the dominating factor $n$! The key idea is to reformulate the \ourslongname{}s in terms of their precision matrix $\bs\Lambda = \bs\Sigma^{-1}$ such that $\bs\gamma\sim\NN\bigl(\bs\mu, \bs\Lambda^{-1}\bigr)$. Consider the general block matrix view as in \cref{eq:block_matrix_regular_cov}:
\begin{equation}
\label{eq:block_matrix_inverse_cov}
    \bs\Sigma^{-1} = \bs\Lambda =
        \begin{bmatrix}
            \bs\Lambda_{aa} & \bs\Lambda_{ab} \\
            \bs\Lambda_{ba} & \bs\Lambda_{bb}
        \end{bmatrix},
\end{equation}
with $\bs\Lambda_{ba} = \bs\Lambda_{ab}^\top$. The conditional mean can now be expressed as
\begin{equation}
\label{eq:conditioning_inverse_cov}
    \bs\mu_{a\vert b} =
        \bs\mu_a -
        \bs\Lambda_{aa}^{-1}
        \bs\Lambda_{ab}
        \bigl(
            \bs\gamma_b - \bs\mu_b
        \bigr).
\end{equation}

\begin{figure}[t]
    \centering
    \begin{minipage}{0.475\textwidth}
        \centering
        \resizebox{\columnwidth}{!}{\includegraphics[width=\linewidth]{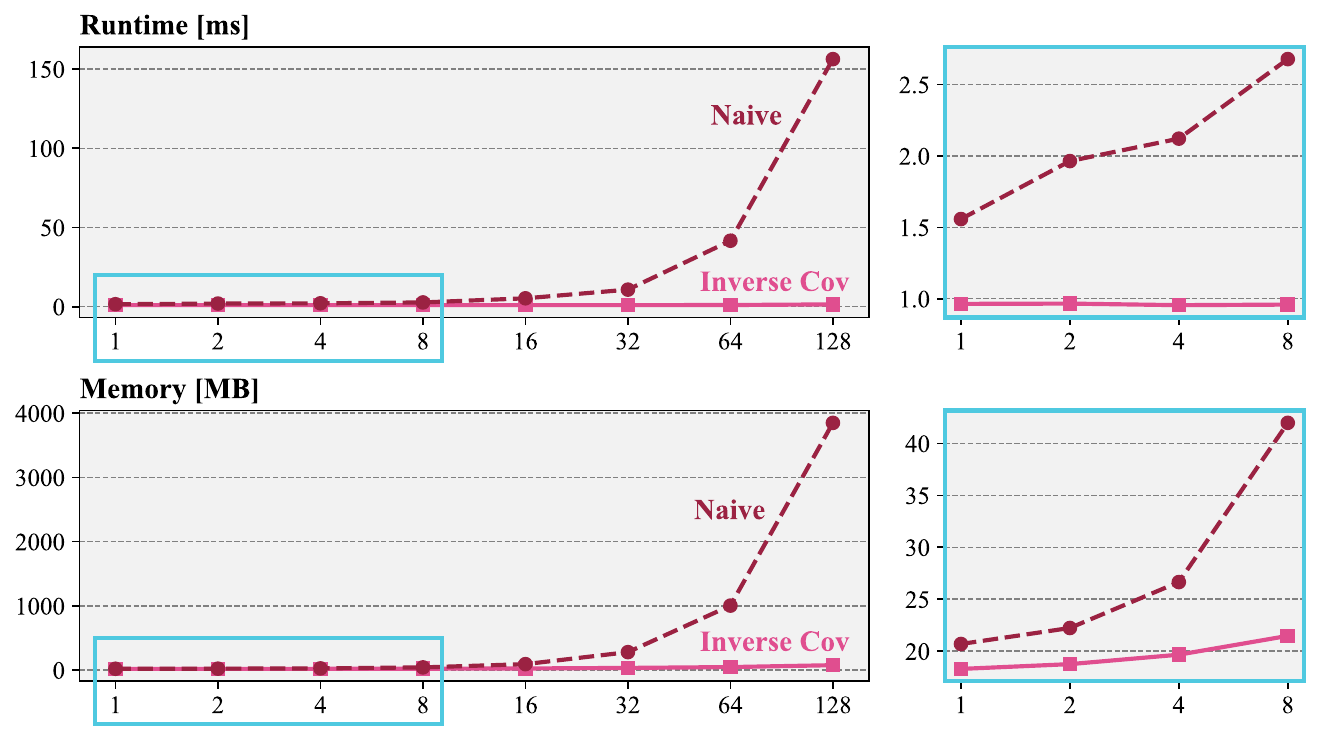}}
    \end{minipage}
    \caption[Naive vs. Inverse Covariance Trick Benchmark]{\textbf{Benchmark Results} on conditioning for ${\sim}15$k \ourslongname{}s with attribute dimension $m = 3$ (\eg, position) and varying latent dimension $n$. We average the measurements across 1000 runs on an NVIDIA GeForce RTX 2080 Ti after an initial warm-up. The benchmark code performs one forward and one backward pass.\label{fig:conditioning_benchmark}}
\end{figure}

And the covariance matrix simply as $\bs\Sigma_{a\vert b} = \bs\Lambda_{aa}^{-1}$. This new formulation only requires storing and inverting the much smaller block $\bs\Lambda_{aa}$ of the precision matrix. While this change seems minor in the derivation, it significantly improves both speed and memory usage. Note that it is not necessary to optimize the full precision matrix $\bs\Lambda$, it suffices only to parameterize $\bs\Lambda_{aa}$ and $\bs\Lambda_{ab}$. This significantly reduces the $\#$parameters from $\mathcal O\bigl((m+n)^2\bigr)$ to $\mathcal O\bigl(m^2 + mn\bigr)$. See \cref{fig:conditioning_benchmark} for a benchmark comparison and the supp. mat. for more details, where we also describe how to further reduce the runtime to $\mathcal O\bigl(m^2n\bigr)$ by exploiting the readily available Cholesky decomposition of $\bs\Lambda_{aa}$.

\subsection{Case Studies}

\oursname{}s can be integrated into existing 3DGS pipelines and improve their quality. To demonstrate this, we inject our conditional \ourslongname{}s into FlashAvatar~\cite{xiang2024flashavatar} and GaussianHeadAvatar \cite{xu2024gaussianheadavatar}. In the following, we briefly describe these two case studies. Please see the supp. mat. for additional details.

\paragraph{Integrating \oursname{}s into FlashAvatar}
\label{ssec:casestudy}
\label{ssec:conditional_multi_gaussian_splatting}
 FlashAvatar deploys a deformation MLP $F_{\theta}$, which maps FLAME~\cite{FLAME:SiggraphAsia2017} expression parameters $\bs\psi$ to per-Gaussian offsets $\Delta\bs\mu_{\bs\psi}$, $\Delta\bs r_{\bs\psi}$, $\Delta\bs s_{\bs\psi}$ for position, rotation, and scale. Note that we adopt the notation from FlashAvatar in this section to ensure consistency. These offsets are then applied to the mesh-attached Gaussians to help model fine details. The deformation MLP further relies on auxiliary input consisting of positional encodings \cite{mildenhall2021nerf} of canonical mesh positions $\bs\mu_{T}$.

\cref{fig:method_overview} illustrates how we replace 3DGS with \oursname{}s. We modify the deformation MLP to output a per-Gaussian latent $\bs z_{\bs\psi}$ instead of offsets. We then compute the conditional distributions $p(\Delta\bs\mu\vert\bs z_{\bs\psi})$, $p(\Delta\bs r\vert\bs z_{\bs\psi})$ and $p(\Delta\bs s\vert\bs z_{\bs\psi})$ for every Gaussian. Recall from \cref{ssec:inverse_covariance_trick} that the conditional means and covariance matrices can be computed efficiently in closed form. The conditional means are then fed to the rest of the pipeline, exactly like the offsets $\Delta\bs\mu_{\bs\psi}$, $\Delta\bs r_{\bs\psi}$, $\Delta\bs s_{\bs\psi}$ from FlashAvatar.

\begin{figure*}
  \centering
  \includegraphics[width=1.0\linewidth]{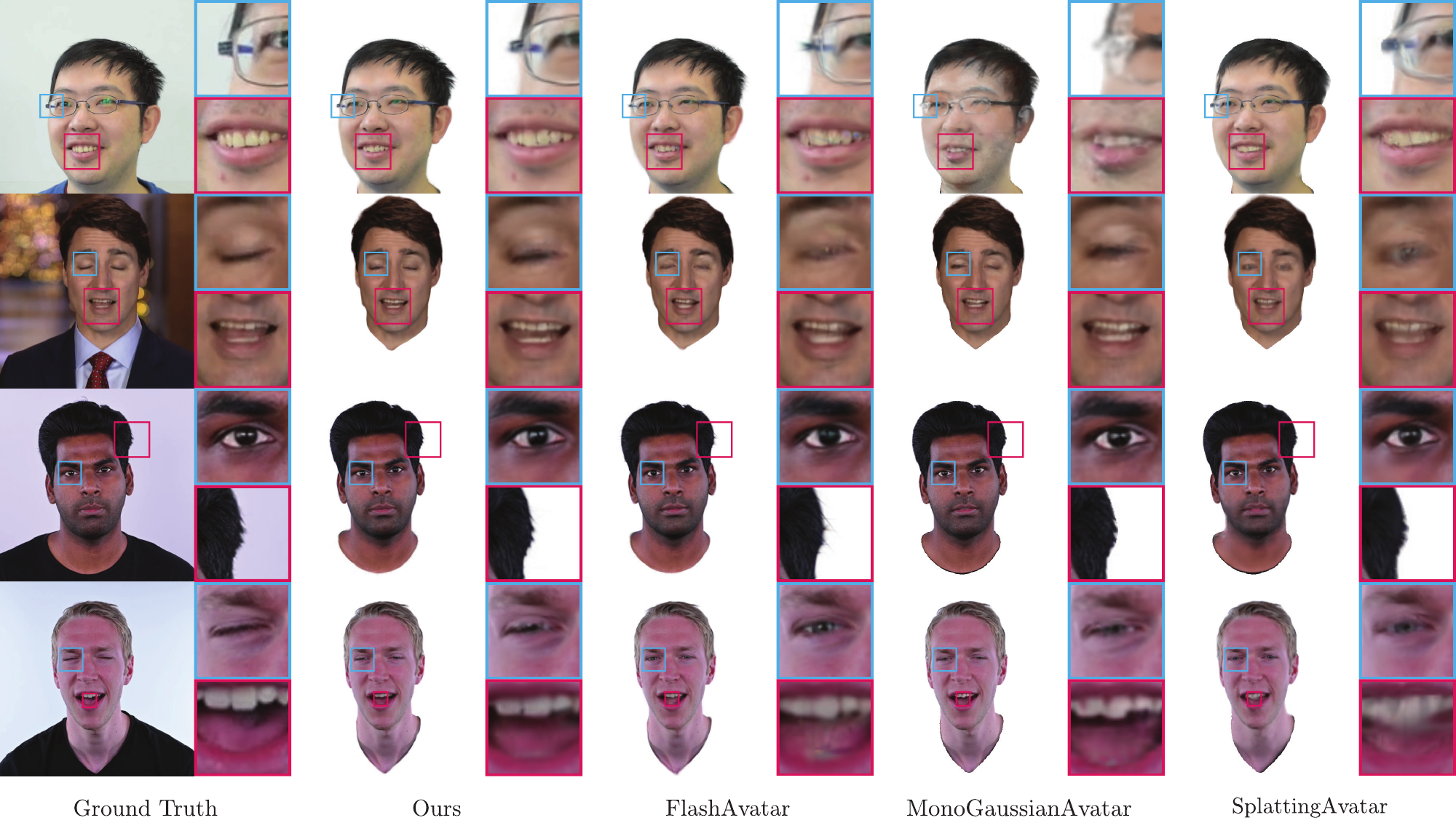}
  \caption{\textbf{Qualitative Comparison} with FlashAvatar \cite{xiang2024flashavatar}, MonoGaussianAvatar \cite{chen2024monogaussianavatar}, and SplattingAvatar \cite{shao2024splattingavatar}. \textbf{Ours} achieves high-quality details for thin structures (glass frames and teeth in the top row), specular reflections (eyes in the third row), and gracefully handles complex deformations (mouth in the second and fourth row).
  \label{fig:qualitative_comparison}}
\end{figure*}

We find that a latent dimensionality of $ n=8$ performs best for face avatars, but even a single latent dimension ($n=1$) improves over the baseline (see the supp. mat. for an ablation).
\cref{fig:teaser,fig:qualitative_comparison} show the effect of HyperGaussians on thin structures like the glass frames, teeth, and specular reflection.

\paragraph{Integrating \oursname{}s into GaussianHeadAvatar}
\label{ssec:casestudy_gha}

To further show the capabilities of \oursname{}s for enhancing existing 3DGS pipelines, we integrate them into the state-of-the-art avatar method for multi-view videos, GaussianHeadAvatar~\cite{xu2024gaussianheadavatar}. GaussianHeadAvatar models the head using expression-conditioned 3D Gaussians with multi-channel colors. More specifically, a dynamic generator takes expression and pose coefficients of a 3DMM and predicts per-Gaussian offsets with respect to a canonical point cloud. The rendered feature image is then passed through a super-resolution module to obtain the final render in \texttt{RGB} space. We modify their architecture to predict latents for our \oursname{}s rather than offsets. Similar to our FlashAvatar modification, we condition the \oursname{}s to obtain regular 3D Gaussians that can be splatted.

Our new representation greatly improves deformation effects, such as wrinkles. Also, it produces sharper reconstructions in general, but particularly in the eyes and glass frames (\cref{fig:teaser,fig:gha-qualitative}). We kindly refer to \citet{xu2024gaussianheadavatar} and the supp. mat. for details on the baseline method.
\noindent

\begin{figure*}
  \centering
  \includegraphics[width=\linewidth]{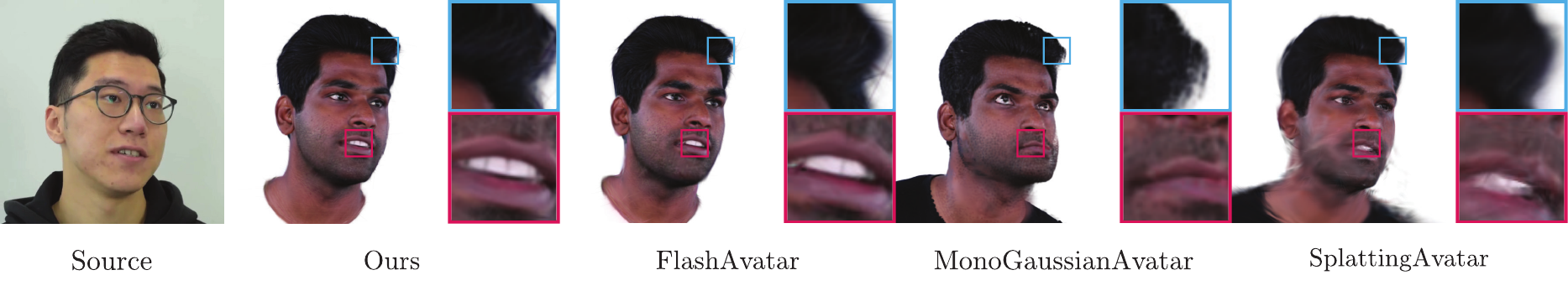}
  \caption{\textbf{Cross-reenactment Comparison} with FlashAvatar \cite{xiang2024flashavatar}, MonoGaussianAvatar \cite{chen2024monogaussianavatar}, and SplattingAvatar \cite{shao2024splattingavatar}. \textbf{Ours} preserves fine details in the teeth and the overall shape of the subject. Please see the supplementary HTML page for more cross-reenactment results.\vspace{-1em}}
  \label{fig:cross_reenactment}
\end{figure*}

\section{Experiments}
\label{sec:experiments}
This section outlines the experimental setting in \cref{ssec:dataset}, compares with the state-of-the-art for face avatars in \cref{ssec:comparison_mono_avatar}, and ablates the key components of \oursname{} in an ablation study in \cref{ssec:ablation_study}.

\subsection{Setting}
\label{ssec:dataset}
\paragraph{Monocular Dataset}
For quantitative results in \cref{tab:quantitative_comparison} (top part) and qualitative comparisons in \cref{fig:qualitative_comparison,fig:cross_reenactment}, we compare on 5 datasets from previous works \cite{zielonka2022insta, zheng2022imavatar, grassal2022neural, Gao2022nerfblendshape, gafni2021dynamic}, consisting of a total of 19 subjects. All videos are subsampled to 25 FPS and resized to $512\times 512$. The videos range from $1$ to $3$ min in length; we train on $2000$ frames and use the last $500$ for testing, following related work \cite{xiang2024flashavatar}.

For preprocessing, we use the same pipeline as FlashAvatar, which consists of MICA~\cite{zielonka2022towards} for FLAME tracking, RVM~\cite{lin2021rvm} for foreground matting, and an off-the-shelf face parser based on BiSeNet~\cite{Yu-ECCV-BiSeNet-2018} for segmentation of the head and neck region, as well as the mouth.

\paragraph{Multi-view Dataset}
We evaluate the effectiveness of our method applied to GaussianHeadAvatar \cite{xu2024gaussianheadavatar} on $10$ subjects from the NeRSemble~\cite{kirschstein2023nersemble} dataset, shown in \cref{tab:quantitative_comparison} (bottom part) and \cref{fig:gha-qualitative}. Following GaussianHeadAvatar, the videos are sampled at 25 FPS, resized, and cropped to $2048\times 2048$. We hold out the FREE sequence for all subjects for testing and train on all remaining sequences. This results in ${\sim}2500$ training frames and ${\sim}300$ test frames, each viewed from $16$ cameras.

For preprocessing, we follow GaussianHeadAvatar and use off-the-shelf tools for background removal \cite{lin2021real} and 2D landmark detection \cite{bulat2017far}. We then track a BFM~\cite{gerig2018morphablebfm2017} model using the landmark annotations.

\paragraph{Training Details}
\label{ssec:implementation_details}
Our case studies on FlashAvatar~\cite{xiang2024flashavatar} and GaussianHeadAvatar~\cite{xu2024gaussianheadavatar} build directly on top of their public PyTorch \cite{paszke2019pytorch} codebases. We inherit all losses and hyperparameters as shown in \cref{ssec:casestudy} and set the learning rate for our \ourslongname{} parameters to $\eta = 10^{-4}$. For FlashAvatar, we train on each video for $30$k iterations, which takes between $15$ and $20$ min on an NVIDIA RTX 4090. For GaussianHeadAvatar, we train on each subject for $600$k iterations, which takes over $2$ days. We publicly release an implementation of \oursname{}s to streamline integration into other codebases.

\begin{table}[b]
    \centering
    \resizebox{\columnwidth}{!}{\begin{tabular}{lccc}
        Method & PSNR ↑ & SSIM$\bigl(10^{-1}\bigr)$ ↑ & LPIPS$\bigl(10^{-2}\bigr)$ ↓ \\
        \hline
        SplattingAvatar~\cite{shao2024splattingavatar} & 28.58 & 9.396 & 9.021 \\
        MonoGaussianAvatar~\cite{chen2024monogaussianavatar} & 29.94 & 9.456 & 6.545 \\
        FlashAvatar (FA)~\cite{xiang2024flashavatar} & 29.43 & 9.466 & 5.107 \\
        \textbf{Ours (FA)} & \textbf{29.99} & \textbf{9.510} & \textbf{4.978} \\
        \hline
        \hline
        GaussianHeadAvatar (GHA)~\cite{xu2024gaussianheadavatar} & 24.10 & 8.819 & 20.273 \\
        \textbf{Ours (GHA)} & \textbf{24.38} & \textbf{8.819} & \textbf{19.768} \\
    \end{tabular}}
    \caption{\textbf{Quantitative comparison with state-of-the-art} digital avatar reconstruction methods from monocular video across 19 subjects from 5 datasets \cite{zielonka2022insta, zheng2022imavatar, grassal2022neural, Gao2022nerfblendshape, gafni2021dynamic} (top) and from multi-view video across 10 subjects from NeRSemble~\cite{kirschstein2023nersemble} (bottom).
    \textbf{Ours (FA)} and \textbf{Ours (GHA)} correspond to FlashAvatar and GaussianHeadAvatar with 8-dimensional \ourslongname{}s without \emph{any other modifications}. Please see \cref{ssec:casestudy,ssec:casestudy_gha} for details.}
    \label{tab:quantitative_comparison}
\end{table}

\subsection{Comparisons}
\label{ssec:comparison_mono_avatar}
\paragraph{Monocular Setting}
We compare with the state-of-the-art for Gaussian-based face avatars: MonoGaussianAvatar \cite{chen2024monogaussianavatar}, SplattingAvatar \cite{shao2024splattingavatar}, and FlashAvatar \cite{xiang2024flashavatar}.
\cref{tab:quantitative_comparison} shows the quantitative comparison against the state-of-the-art on the commonly used metrics PSNR, SSIM, and LPIPS~\cite{zhang2018unreasonable} with a VGG~\cite{simonyan2014very} backbone.

\cref{fig:qualitative_comparison} shows a qualitative comparison for self-reenactment.
SplattingAvatar \cite{shao2024splattingavatar} is not designed to represent expression-dependent appearance effects. This leads to unrealistic renderings in reflective regions, such as eyes and glasses, and to artifacts in areas with strong deformations, such as the mouth and eyes. Note how it fails to close the eyelid in the second row.
MonoGaussianAvatar \cite{chen2024monogaussianavatar} predicts expression-dependent offsets but struggles with specular reflections.
FlashAvatar uses an MLP to predict expression-dependent Gaussian parameter offsets. Their output lacks detail for thin structures, such as the glass frames or gaps between teeth, and it does not handle specular reflections on the eyes and glasses well. \textbf{Ours} handles such high-frequency details more gracefully. 
It is crucial to keep in mind that the \emph{only difference} between \textbf{Ours} and FlashAvatar is the Gaussian representation. Without any other changes or hyperparameter tuning, \oursname{}s are capable of rendering more accurate specular reflections and thin structures in the mouth, eyes, and glass frames. 

We qualitatively compare cross-reenactment in \cref{fig:cross_reenactment}. Note the distorted mouth and blurry hair in SplattingAvatar. MonoGaussianAvatar renders a more realistic face but exhibits unrealistic jaw deformations. FlashAvatar can render the correct expression, but the teeth and hair appear blurry. \textbf{Ours} produces higher-quality renders showing individual teeth and sharper hair.
This can also be observed in a side-by-side comparison for training convergence (see supp. mat.).

Finally, it is important to mention that MonoGaussianAvatar is substantially heavier than FlashAvatar and \textbf{Ours}. MonoGaussianAvatar requires over $12$h of training time and uses over $100$k Gaussians. FlashAvatar and \textbf{Ours} are much lighter, with only ${\sim}15$k Gaussians and less than $20$min of training time. Despite this, our \oursname{}s outperforms them both. We encourage readers to visit the supplementary HTML page to see animated results.

\begin{figure}[t]
    \centering
    \includegraphics[width=\linewidth]{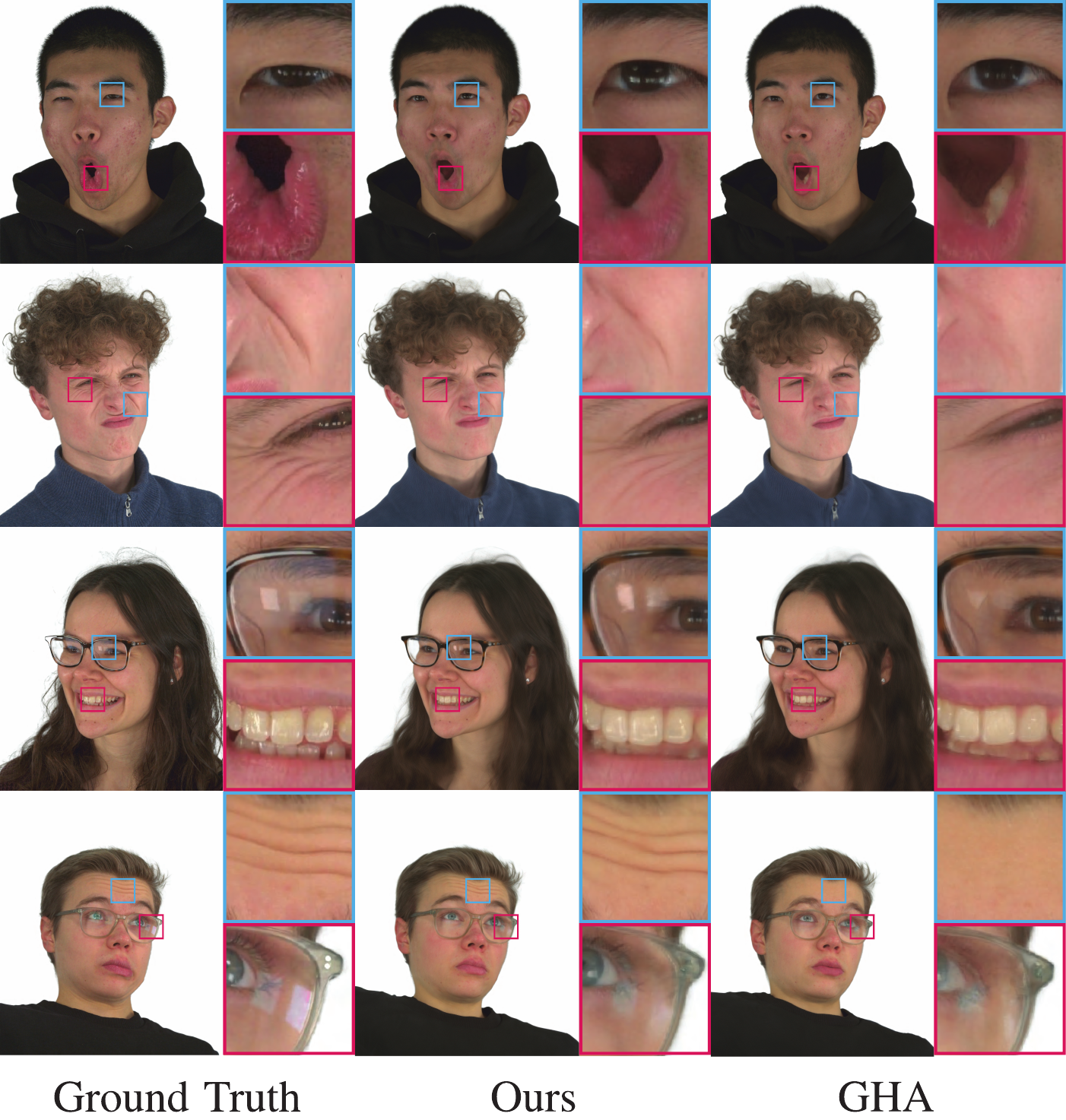}
    \caption{Enhancing GaussianHeadAvatars \cite{xu2024gaussianheadavatar} with \oursname{}s boosts high-frequency details like wrinkles and reflections in the eyes, glasses, and teeth. See \cref{tab:quantitative_comparison} for metrics.}
    \label{fig:gha-qualitative}
    \vspace{-1em}
\end{figure}

\paragraph{Multi-view Setting}
We further evaluate HyperGaussians on multi-view avatar reconstruction. By replacing 3D Gaussians with our \oursname{}s, we significantly improve the visual quality of GaussianHeadAvatar (\cref{fig:gha-qualitative}). Our more expressive representation enables the model to faithfully reconstruct challenging nonlinear effects. This is evident in rows 1, 3, and 4, where our method improves the appearance of specular highlights on eyes, teeth, and glasses. Moreover, the rich local context of \oursname{}s allows for accurate modeling of complex skin deformations, which the baseline failed to capture entirely (rows 2 and 4).

Note that, thanks to the inverse covariance trick, the \oursname{}s integration results in very little overhead. Specifically, the training time increased by only $30$min ($1\%$), which is negligible considering the total of over $2$d.

\subsection{Ablation Study}
\label{ssec:ablation_study}

We conduct ablation studies in the monocular setting (FlashAvatar \cite{xiang2024flashavatar}, \cref{ssec:casestudy}). We first ablate alternative means of increasing expressiveness by adding parameters to other parts of the model, and then study the computational performance gains in speed and memory when employing the inverse covariance trick (\cref{ssec:inverse_covariance_trick}). The supp. mat. contains ablations for different latent dimensionalities.

\paragraph{Scaling the Model}
The quality improvements provided by HyperGaussians cannot be matched by scaling other parts of the model. To demonstrate this, we increase the width and depth of the deformation MLP $F_\Theta$ (\cref{fig:method_overview}). Increasing the MLP depth deteriorates perceptual quality (LPIPS \cite{zhang2018unreasonable}, lower is better) by $15\%$ (from $0.0498$ for HyperGaussians to $0.0572$ for a deeper MLP with the same number of parameters) and reduces the rendering framerate by $47\%$ from 300 FPS to 158 FPS. Increasing MLP width deteriorates perceptual quality by $3\%$ (from $0.0498$ for HyperGaussians to $0.0512$ for a wider MLP with the same number of parameters) and reduces the rendering framerate by $41\%$ from 300 FPS to 178 FPS. Please refer to the supp. mat. to see a detailed ablation for different configurations.

\paragraph{Inverse Covariance Trick}
\label{ssec:ablation_inverse_covariance_trick}
As shown in \cref{ssec:inverse_covariance_trick}, a na\"ive implementation of the conditioning in \cref{eq:conditioning_regular_cov}, as in NDGS~\cite{diolatzis2024n}, is very inefficient for large latent codes $\bs\gamma_b$, due to the large matrix $\bs\Sigma_{bb}\in\R^{n\times n}$. After applying the inverse covariance trick (\cref{ssec:inverse_covariance_trick}), we reduce the problem to inverting the small matrix $\bs\Lambda_{aa}\in\R^{m\times m}$ ($m\in\{3, 4\}$).
\cref{fig:conditioning_benchmark} shows an empirical ablation for latent dimensionalities $n$ between $1$ and $128$. In our case study on FlashAvatar~\cite{xiang2024flashavatar} (\cref{ssec:casestudy}), the inverse covariance trick improves speed by $150\%$ for a small latent $n=8$ and by $15,000\%$ for a large latent $n=128$. Not only does the inverse covariance matrix improve speed, but it also reduces memory usage. For a small latent $n=8$, the na\"ive implementation uses $42$ MB, whereas the inverse covariance trick reduces this to $22$ MB, a reduction of $48\%$. For $n=128$, it is over $90\%$.
\section{Conclusion and Discussion}
\label{sec:conclusion}
In this paper, we study how 3DGS can be made more expressive for monocular and multi-view face avatars. The result, \oursname{}s, is a novel extension to 3D Gaussians that offers improved expressivity and rendering quality, excelling in high-frequency details. We motivate our extension by taking a Bayesian perspective and showing that \oursname{}s perform MAP estimation over conditional distributions across 3DGS scenes.
Our evaluations on 29 subjects across 6 datasets outperform the state-of-the-art by simply \emph{plugging} our proposed \oursname{}s into FlashAvatar and GaussianHeadAvatar, \emph{without any other modifications}.
As a limitation, high-dimensional variants of 3DGS require significantly more compute and memory, which has prevented widespread adoption of NDGS. Our proposed \emph{inverse covariance trick} substantially reduces these requirements and opens up future research directions that could apply \oursname{}s beyond human faces, \emph{i.e.}, in complex settings that require higher latent dimensions. In conclusion, \oursname{}s show great promise for improving high-frequency details, bringing the field a step closer to photorealistic and fast face avatars.

\paragraph*{Acknowledgments}
We thank Jenny Schmalfuss, Seonwook Park, Lixin Xue, Zetong Zhang, Chengwei Zheng, Egor Zakharov for fruitful discussions and proofreading, and Yufeng Zheng for help with dataset preprocessing.
{
    \small
    \bibliographystyle{ieeenat_fullname}
    \bibliography{main}
}

\clearpage
\appendix
\section*{Supplementary Material}
This supplement contains more details and derivations in \cref{suppsec:details,suppsec:gha_details}, supplementary results in \cref{suppsec:results}, and discusses the potential societal impact of this work in \cref{suppsec:impact}.

\section{Positioning with respect to Related Works}
\label{suppsec:positioning}

We provide an overview of the main differences between \oursname{}s and the most closely related works on Gaussian avatars in \cref{tab:relatedwork}. We partition the methods into 3 categories: optimization-based, feed-forward, and pseudo ground-truth. Since the latter two categories require substantial computational resources for pre-training, we mainly focus on optimization-based techniques in our experiments.

\begin{table*}
    \centering
    \begin{tabular}{l|cccc}
      Method & Representation & Local context & Dynamic context & Resource requirements \\
      \hline
      GaussianAvatars \cite{qian2024gaussianavatars} & 3DGS & $\times$ & $\times$ & Low \\
      SurFHead \cite{lee2025surfhead} & 2DGS & $\times$ & $\times$ & Low \\
      SplattingAvatar \cite{shao2024splattingavatar} & 3DGS & $\checkmark$ & $\times$ & Low \\
      MonoGaussianAvatar \cite{chen2024monogaussianavatar} & 3DGS & $\times$ & $\checkmark$ & Moderate \\
      FlashAvatar (FA) \cite{xiang2024flashavatar} & 3DGS & $\times$ & $\checkmark$ & Low \\
      GaussianHeadAvatar (GHA) \cite{xu2024gaussianheadavatar} & 3DGS & $\times$ & $\checkmark$ & Moderate \\
      \hline
      GAGAvatar \cite{chu2024generalizable} & FF 3DGS & $\checkmark$ & $\checkmark$ & High \\
      LAM \cite{he2025lam} & FF 3DGS & $\checkmark$ & $\times$ & High \\
      FastAvatar \cite{liang2025fastavatar} & FF 3DGS & $\checkmark$ & $\times$ & High \\
      Avat3r \cite{tang2025gaf} & FF 3DGS & $\checkmark$ & $\checkmark$ & High \\
      \hline
      GAF \cite{tang2025gaf} & MVDiffusion 3DGS & $\times$ & $\times$ & Very High \\
      CAP4D \cite{taubner2025cap4d} & MVDiffusion 3DGS & $\times$ & $\times$ & Very High \\
      \hline
      \hline
      \textbf{Ours (FA)} & HGS & $\checkmark$ & $\checkmark$ & Low \\
      \textbf{Ours (GHA)} & HGS & $\checkmark$ & $\checkmark$ & Moderate
    \end{tabular}
    \caption{\textbf{Differences to the most closely related works.} GaussianAvatars \cite{qian2024gaussianavatars} and SuRFHead \cite{lee2025surfhead} deform 3D Gaussians based on an underlying FLAME mesh \cite{FLAME:SiggraphAsia2017} without local embeddings or dynamic inputs like facial expressions. SplattingAvatar \cite{shao2024splattingavatar} optimizes local embeddings, but the Gaussian properties are not dependent on expressions or pose. MonoGaussianAvatar \cite{chen2024monogaussianavatar}, FlashAvatar \cite{xiang2024flashavatar}, and GaussianHeadAvatar \cite{xu2024gaussianheadavatar} predict expression-dependent offsets to the Gaussian properties, but their lack of local context leads to blurry or distorted results, see comparison in the main paper. Our proposed representation (HGS) attaches high-dimensional Gaussians to the mesh and optimizes learnable local embeddings for modulating the Gaussian properties based on expressions. A more recent line of work deploys large-scale generative models to either directly regress 3D Gaussian parameters or augment the training signal during optimization. GAGAvatar~\cite{chu2024generalizable} and LAM~\cite{he2025lam} derive 3D Gaussians from DINOv2~\cite{oquab2023dinov2} features. FastAvatar~\cite{liang2025fastavatar} and Avat3r~\cite{kirschstein2025avat3r} train an encoder-decoder model from scratch, where Avat3r additionally uses position maps from DUSt3R~\cite{wang2024dust3r} and feature maps from Sapiens~\cite{khirodkar2024sapiens}. GAF~\cite{tang2025gaf} and CAP4D~\cite{taubner2025cap4d} train multi-view diffusion models \cite{poole2022dreamfusion,shi2023mvdream,gao2024cat3d} to generate pseudo ground-truth images for 3DGS optimization. These feed-forward and pseudo GT methods show promising results in sparse-view settings, their primary focus. However, they lose the ability to leverage all information when more views are available. The typical range is 1 to 4 images per forward pass. CAP4D proposes a sophisticated inference scheme to address this issue. However, it still takes 12 GPU hours to produce a single avatar. Traditional optimization-based techniques still achieve superior performance at significantly lower computational effort in such cases.}
    \label{tab:relatedwork}
\end{table*}

\section{\oursname{} Details}
\label{suppsec:details}
This section provides more details about the formulation of \oursname{}s.

\subsection{Parameterization}
\label{suppssec:parameteriztion}
Each \oursname{} consists of a high-dimensional mean $\bs\mu$ and covariance matrix $\bs\Sigma$ (or precision matrix $\bs\Lambda$ when applying the \emph{inverse covariance trick}, \cref{supssec:inversecov}) with optimizable parameters. We parameterize the mean $\bs\mu$ directly and decompose the covariance $\bs\Sigma$ into its Cholesky factor $\bs L$ \cite{diolatzis2024n}, such that $\bs\Sigma = \bs L\bs L^\top$, where $\bs L$ is a lower triangular matrix with positive diagonal entries. To ensure the uniqueness of the factorization, we apply an exponential activation function, $\bs L_{i, i} \gets \exp \bs L_{i, i}$, to the diagonal entries of the parameter matrix. We further show in \cref{supssec:inversecov} that it is not necessary to parameterize the entire Cholesky factor $\bs L$ when working with the inverse covariance trick (\cref{suppeq:conditioning_inverse_cov,eq:conditioning_inverse_cov2}).

\subsection{Splatting}
\label{suppssec:splatting}
This paragraph explains how we construct the 3D covariance matrix for splatting the conditioned Gaussians. We build on top of the same decomposition of $\bs\Sigma_{\mathrm{3D}}$ into rotation and scaling as in vanilla 3DGS, but replace the static terms $\bs R$ and $\bs S$ by their conditional analogues. More formally,
\begin{align}
    \bs\Sigma_{\mathrm{3D}} = \bs R(\bar{\bs q})\bar{\bs S}\bar{\bs S}^\top\bs R(\bar{\bs q})^\top,
\end{align}

where $\bs R(\cdot)$ denotes quaternion to rotation matrix conversion, $\bar{\bs q} = \mathbb{E}[\mathcal A_{\bs q}\vert\bs z]$, and $\bar{\bs S} = \mathrm{diag}\bigl(\mathbb{E}[\mathcal A_{\bs s}\vert\bs z]\bigr)$. Moreover, $\bs z$ is the latent code and $\mathcal A$ stands for \emph{attribute} (see Eq. (4) in the main paper).

This is an important distinction to NDGS~\cite{diolatzis2024n}, which directly splats using the conditional covariance matrix $\bs\Sigma_{\mathrm{3D}} = \mathrm{Cov}[\mathcal{A}_{\bs\mu_{\mathrm{3D}}}\vert\bs z]$. NDGS has no degrees of freedom for the conditional orientation of the 3D Gaussians, \emph{i.e.}, the 3D Gaussians cannot conditionally rotate.

\begin{figure}
  \centering
  \includegraphics[width=\linewidth]{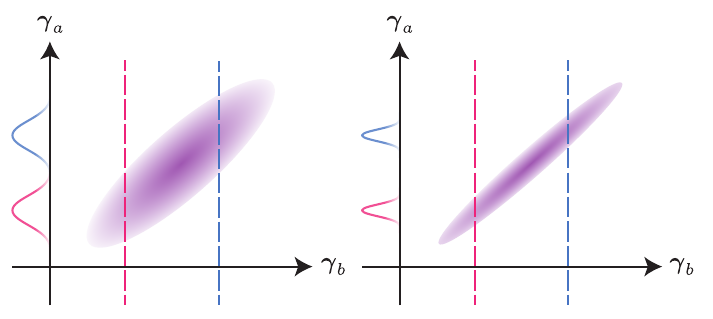}
  \caption{\textbf{Geometric interpretation of Gaussian conditioning} on two examples with large (left) and small (right) uncertainty at different realizations of $\bs\gamma_b$. As a result, the conditional mean shifts, while the conditional covariance is the same for both slices.}
  \label{suppfig:slicing}
\end{figure}

To illustrate this, note that conditioning can be interpreted geometrically as taking an $m$-dimensional slice through the multivariate Gaussian with $(m+n)$ total dimensions. \cref{suppfig:slicing} shows two examples for $m = n = 1$. Notice that the conditional covariance matrix has the same shape regardless of where the slice is taken (Eq. (3) in the main paper).

We instead use $\mathrm{Cov}[\mathcal{A}_{\bs\mu_{\mathrm{3D}}}\vert\bs z]$, $\mathrm{Cov}[\mathcal{A}_{\bs q}\vert\bs z]$, and $\mathrm{Cov}[\mathcal{A}_{\bs s}\vert\bs z]$ for visualizing uncertainties (see \cref{suppfig:uncertainty}).

\subsection{Derivation of the Inverse Covariance Trick}
\label{supssec:inversecov}
We explain in the main paper that a na\"ive implementation of the conditioning is very inefficient for large latent codes $\bs\gamma_b$. Here, we provide a more detailed derivation of the inverse covariance trick.

We reformulate our \ourslongname{}s in terms of their precision matrix $\bs\Lambda = \bs\Sigma^{-1}$ such that $\bs\gamma\sim\NN\bigl(\bs\mu, \bs\Lambda^{-1}\bigr)$. We consider the following block matrix view:
\begin{equation}
\label{suppeq:block_matrix_inverse_cov}
    \bs\Sigma^{-1} = \bs\Lambda =
        \begin{bmatrix}
            \bs\Lambda_{aa} & \bs\Lambda_{ab} \\
            \bs\Lambda_{ba} & \bs\Lambda_{bb}
        \end{bmatrix}
\end{equation}
with $\bs\Lambda_{ba} = \bs\Lambda_{ab}^\top$.

As the inverse of $\bs\Sigma$, $\bs\Lambda$ inherits symmetry, \cref{eq:precision_matrix_symmetric}, as well as positive definiteness since its eigenvalues are the reciprocal of the eigenvalues of $\bs\Sigma$, and therefore all positive:
\begin{equation}
\label{eq:precision_matrix_symmetric}
\bs\Lambda^\top = \bigl(\bs\Sigma^{-1}\bigr)^\top = \bigl(\bs\Sigma^\top\bigr)^{-1} = \bs\Sigma^{-1} = \bs\Lambda.
\end{equation}
This is important, as it allows us to reuse the same parameterization that we described in \cref{suppssec:parameteriztion} to represent $\bs\Lambda = \bs L\bs L^\top$. Conveniently, we also get the Cholesky factor $\bs L_{11}$ of $\bs\Lambda_{aa}$ as a side product:
\begin{equation}
\begin{split}
\label{eq:block_matrix_cholesky_inverse_cov}
    \begin{bmatrix}
        \bs\Lambda_{aa} & \bs\Lambda_{ab} \\
        \bs\Lambda_{ba} & \bs\Lambda_{bb}
    \end{bmatrix}
    &=
    \begin{bmatrix}
        \bs L_{11} &
        \bs 0 \\
        \bs L_{21} &
        \bs L_{22}
    \end{bmatrix}
    \begin{bmatrix}
        \bs L_{11}^\top &
        \bs L_{21}^\top \\
        \bs 0 &
        \bs L_{22}^\top
    \end{bmatrix}
    \\ &=
    \begin{bmatrix}
        \bs L_{11} \bs L_{11}^\top &
        \bs L_{11} \bs L_{21}^\top \\
        \bs L_{21} \bs L_{11}^\top &
        \bs L_{21} \bs L_{21}^\top +
        \bs L_{22} \bs L_{22}^\top
    \end{bmatrix}.
\end{split}
\end{equation}

With this new formulation, the conditional mean and covariance matrix can be expressed as
\begin{equation}
\begin{split}
\label{suppeq:conditioning_inverse_cov}
    \bs\mu_{a\vert b} &=
        \bs\mu_a -
        \bs\Lambda_{aa}^{-1}
        \bs\Lambda_{ab}
        \bigl(
            \bs\gamma_b - \bs\mu_b
        \bigr)
    \\
    \bs\Sigma_{a\vert b} &= \bs\Lambda_{aa}^{-1},
\end{split}
\end{equation}
where the term $\bs\Lambda_{aa}^{-1}\bs\Lambda_{ab}$ can be further broken down into
\begin{equation}
\label{eq:conditioning_inverse_cov2}
    \bs\Lambda_{aa}^{-1}
    \bs\Lambda_{ab}
    =
    \bigl(\bs L_{11}\bs L_{11}^\top\bigr)^{-1}\bs L_{11}\bs L_{21}^\top
    =
    \bs L_{11}^{-\top}\bs L_{21}^\top.
\end{equation}
Note that products with $\bs L_{11}^{-\top}$ can be evaluated in an efficient and numerically stable manner since $\bs L_{11}$ is a triangular matrix of small size, which is independent of the latent dimension. These equations also show that it is sufficient only to parameterize $\bs L_{11}$ and $\bs L_{21}$ for modeling the conditional distribution. This reduces the number of parameters from $\mathcal{O}\bigl((m + n)^2\bigr)$ to $\mathcal{O}\bigl(m^2 + mn\bigr)$ compared to the full parameterization. Please see the main paper for a benchmark comparison between the na\"ive implementation and the one applying the inverse covariance trick.

\subsection{Derivation of Uncertainty}
\label{suppssec:uncertainty_quantification}
We observe an interesting property about \ourslongname{}s, which arises naturally from their Bayesian interpretation. \ourslongname{}s are at their core multivariate Gaussian distributions. Their conditional covariance matrices indicate the variance of each Gaussian across the different expressions of the training subject and can be intuitively interpreted as uncertainty.

More formally defined, we have
\begin{equation}
\begin{split}
\label{eq:uncertainty_def}
\begin{aligned}
\sigma
:=& \log\det\bs\Sigma_{a\vert b} \\
=& -\log\det\bs\Lambda_{aa} && (1) \\
=& -2\log\det\bs L_{11} && (2) \\
=& -2\tr\log\bs L_{11}, && (3)
\end{aligned}
\end{split}
\end{equation}
where we used $\det\bs\Sigma_{a\vert b} = \det\bs\Lambda_{aa}^{-1} = \bigl(\det\bs\Lambda_{aa}\bigr)^{-1}$ in step $(1)$, $\det\bs\Lambda_{aa} = \det\bs L_{11}\bs L_{11}^\top = \bigl(\det\bs L_{11}\bigr)^2$ in step $(2)$, and $\det\bs L_{11} = \prod_{i=1}^m (\bs L_{11})_{i, i}$ in step $(3)$. The $\log$ in step $(3)$ is applied element-wise.

\begin{figure}
  \centering
  \includegraphics[width=\linewidth]{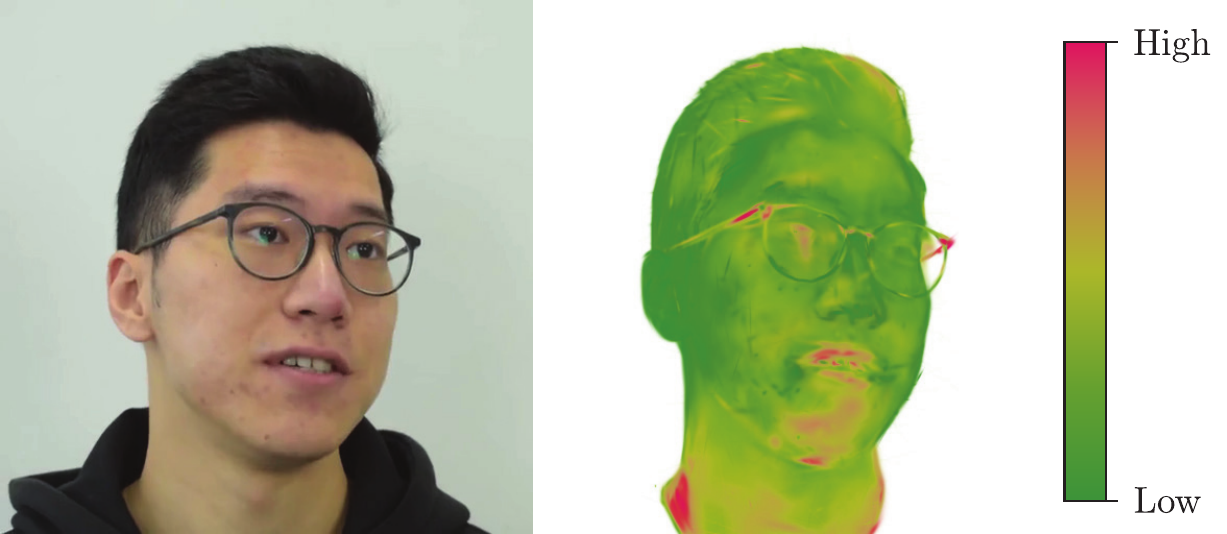}
  \caption{\textbf{Uncertainty quantification} on one of the training subjects. \colorbox[HTML]{66B56B}{Green} denotes low uncertainty, while \colorbox[HTML]{E66084}{Red} denotes high uncertainty. Note that the semantic structure arises purely from the probabilistic formulation without additional supervision.}
  \label{suppfig:uncertainty}
\end{figure}

Again, the inverse covariance trick (\cref{supssec:inversecov}) is key to efficiently compute this quantity. These values are summed up across the conditional distributions for all Gaussian attributes. To render these uncertainties, we further apply a sigmoid function and map the values to colors. This agreement between the uncertainty estimates and what would intuitively be considered difficult regions emerges without explicit supervision. We demonstrate an example of this effect in \cref{suppfig:uncertainty}. Note that the foreground mask in the bottom-left neck area was unstable throughout the video, leading to high uncertainty despite being rigid. Moreover, the uncertainty correctly captures the variability of the lips and specular reflections on glasses. We further observe that, while rigid, the glass frames exhibit significant displacements due to underlying mesh deformations.

\section{GaussianHeadAvatar Details}
\label{suppsec:gha_details}

GaussianHeadAvatar (GHA) \cite{xu2024gaussianheadavatar} represents a scene with $N$ Gaussians, which have position $X$, multi-channel color $C$, rotation $Q$, scale $S$ and opacity $A$. The way GHA models expression-dependent effects is by maintaining a canonical set of Gaussians $\{X_0, F_0, Q_0, S_0, A_0\}$, where $F_0$ is a per-point feature vector, and training an MLP-based expression conditioning dynamic generator $\Phi$, which predicts dynamic changes with respect to the canonical model. More specifically, given expression $\theta$ and head pose $\beta$, they compute
\begin{equation}
\label{suppeq:gha_deformation_mlp}
    \{X, C, Q, S, A\} = \Phi(X_0, F_0, Q_0, S_0, A_0;\theta, \beta)
\end{equation}

in order to obtain a deformed set of Gaussians.

For our \oursname{}s integration, we modify the pipeline such that the MLP outputs per-Gaussian latent codes for each attribute, which are then used to condition the \oursname{}s. Formally,
\begin{equation}
\label{suppeq:gha_hgs_deformation_mlp}
    \{\bs z_X, \bs z_C, \bs z_Q, \bs z_S, \bs z_A\} = \Phi(X_0, F_0, Q_0, S_0, A_0;\theta, \beta)
\end{equation}

are the expression-dependent latents, which are then used for computing the conditional means $\mathbb{E}[\mathcal A\vert \bs z]$ that are fed to the remainder of the pipeline, just like the MLP offsets in the original method, analogous to our FlashAvatar integration.

\section{Supplementary Experiments}
\label{suppsec:results}
\paragraph{NeRSemble Details} For our multi-view setting, we used 10 subjects from the NeRSemble \cite{kirschstein2023nersemble} dataset, with the following IDs: 017, 018, 024, 031, 033, 036, 037, 129, 141, 144. We select all sequences labelled with \texttt{EMO}, \texttt{EXP} (excluding tongue sequences), \texttt{SEN} for training, and the \texttt{FREE} sequences for testing.
\subsection{Qualitative Results}
\begin{figure}[H]
    \centering
    \includegraphics[width=\linewidth]{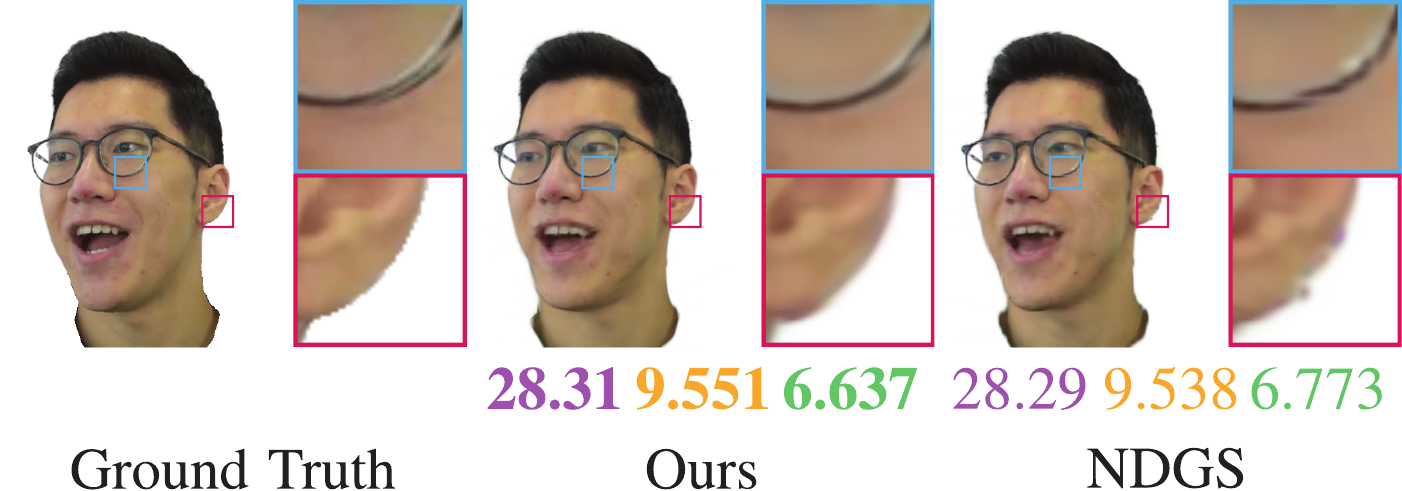}
    \caption{\textbf{Comparison with NDGS} integrated into FlashAvatar. The limited degrees of freedom in NDGS lead to misalignments of thin structures and edges. The numbers show \textcolor{psnrcolor}{PSNR}, \textcolor{ssimcolor}{SSIM ($10^{-1}$)}, and \textcolor{lpipscolor}{LPIPS ($10^{-2}$)}.}
    \label{fig:ndg-qualitative}
\end{figure}

We show video results for self- and cross-reenactment on the supplementary HTML page. In addition, \cref{fig:ndg-qualitative,suppfig:gha-ndg-qualitative} show additional results against NDGS~\cite{diolatzis2024n} after integration into the baseline methods, similar to \oursname{}s. Moreover, \cref{suppfig:ablation} provides qualitative results for varying latent dimensionalities.

\subsection{Comparison with NDGS}

\begin{figure*}
    \centering
    \includegraphics[width=\linewidth]{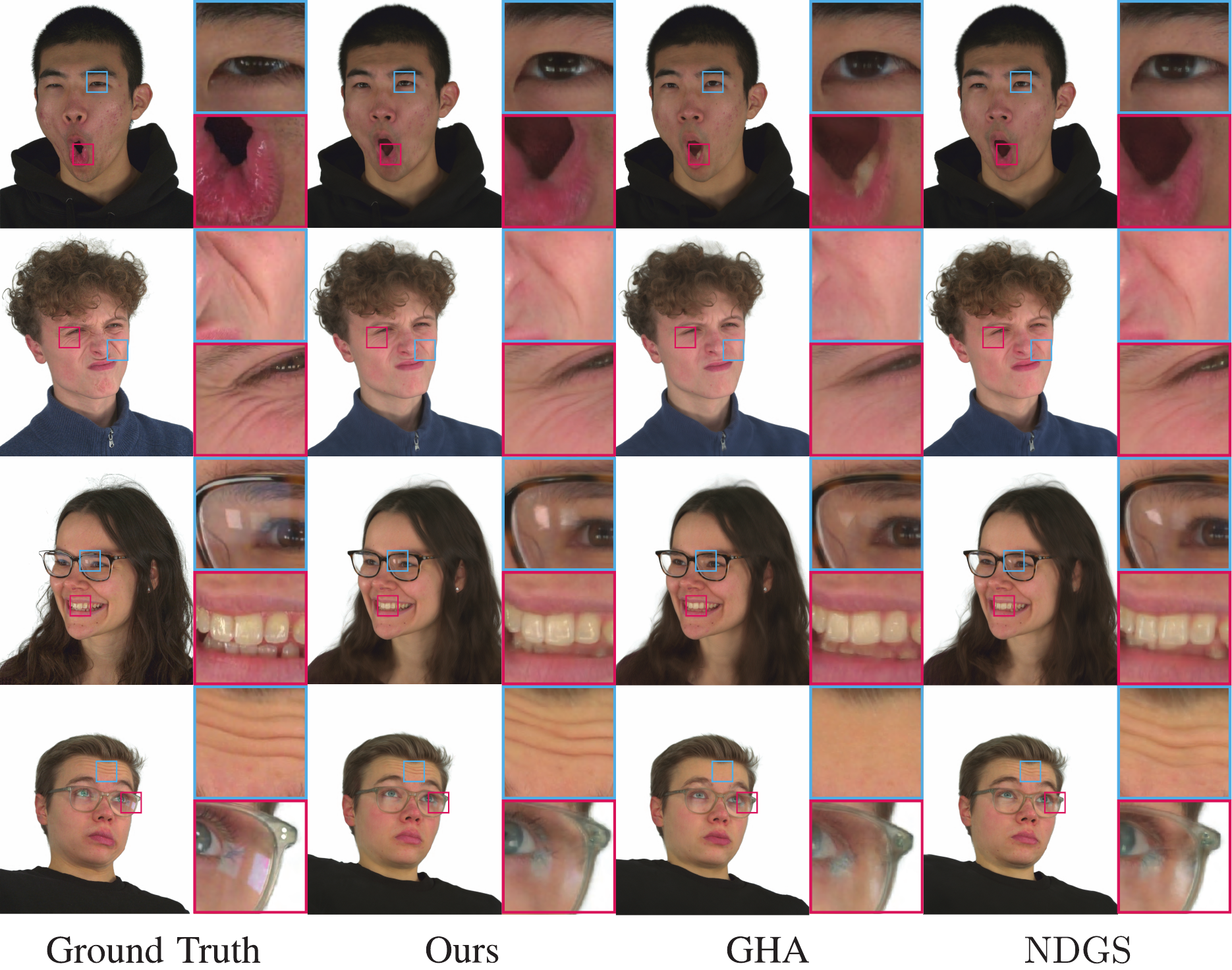}
    \caption{\textbf{Qualitative Comparison} with GHA and NDGS on subjects from the NeRSemble \cite{kirschstein2023nersemble} dataset. While NDGS helps improve skin deformations, the resulting wrinkles are less pronounced (rows 2 and 4 show blurrier, washed-out results). Moreover, NDGS lacks visual fidelity for specular highlights (row 3) and struggles with accurate geometric alignment (teeth artifacts in row 1 and downward displacement of the glass frames in row 4).}
    \label{suppfig:gha-ndg-qualitative}
\end{figure*}

Our approach differs from NDGS~\cite{diolatzis2024n} in its mathematical formulation and capabilities. While NDGS uses the conditional covariance matrix, which is \emph{independent} of the latent code, to represent the size and shape directly, our \oursname{}s apply multivariate Gaussians on each attribute where the \emph{conditional means dynamically adapt the location, scale, and orientation} of the derived 3D Gaussians \emph{in response to} the latent code. This crucial difference gives \oursname{}s the necessary degrees of freedom to model thin geometry and complex deformations with higher accuracy. \cref{fig:ndg-qualitative} shows our method eliminates the artifacts visible in NDGS (particularly on glass frames) and produces substantially more precise geometry at curved boundaries. We further demonstrate the benefits of \oursname{}s over NDSG by integrating them into GaussianHeadAvatar \cite{xu2024gaussianheadavatar} in a similar fashion. Again, we can observe in \cref{suppfig:gha-ndg-qualitative} that NDGS produces inferior reconstructions compared to \oursname{}s. More specifically, it lacks detail for specular reflections on the glasses and teeth (row 3), and it fails to accurately position the glasses at the correct vertical location (a downward shift in row 4 due to surrounding skin deformations). Moreover, the bottom-right wrinkles in row 4 disappear when the expression latent deviates too far from the mean, since NDGS uses the joint probability density to modulate opacity. These observations can also be made quantitatively in \cref{supptab:quantitative_comparison}.

\begin{table}[h]
    \centering
    \resizebox{\columnwidth}{!}{\begin{tabular}{lccc}
        Method & PSNR ↑ & SSIM$\bigl(10^{-1}\bigr)$ ↑ & LPIPS$\bigl(10^{-2}\bigr)$ ↓ \\
        \hline
        GaussianHeadAvatar (GHA)~\cite{xu2024gaussianheadavatar} & 24.10 & 8.819 & 20.273 \\
        NDGS (GHA)~\cite{diolatzis2024n} & 24.22 & \textbf{8.820} & 20.235 \\
        \textbf{Ours (GHA)} & \textbf{24.38} & 8.819 & \textbf{19.768} \\
    \end{tabular}}
    \caption{\textbf{Quantitative comparison} with GHA and NDGS in the multi-view setting. Notice that NDGS is unable to match the improvements of \oursname{}s in terms of PSNR and LPIPS, and performs only marginally better on SSIM. This highlights the limited capabilities of NDGS due to its reduced degrees of freedom.}
    \label{supptab:quantitative_comparison}
\end{table}

\subsection{Ablation Study}
\begin{figure*}
  \centering
  \includegraphics[width=1.0\linewidth]{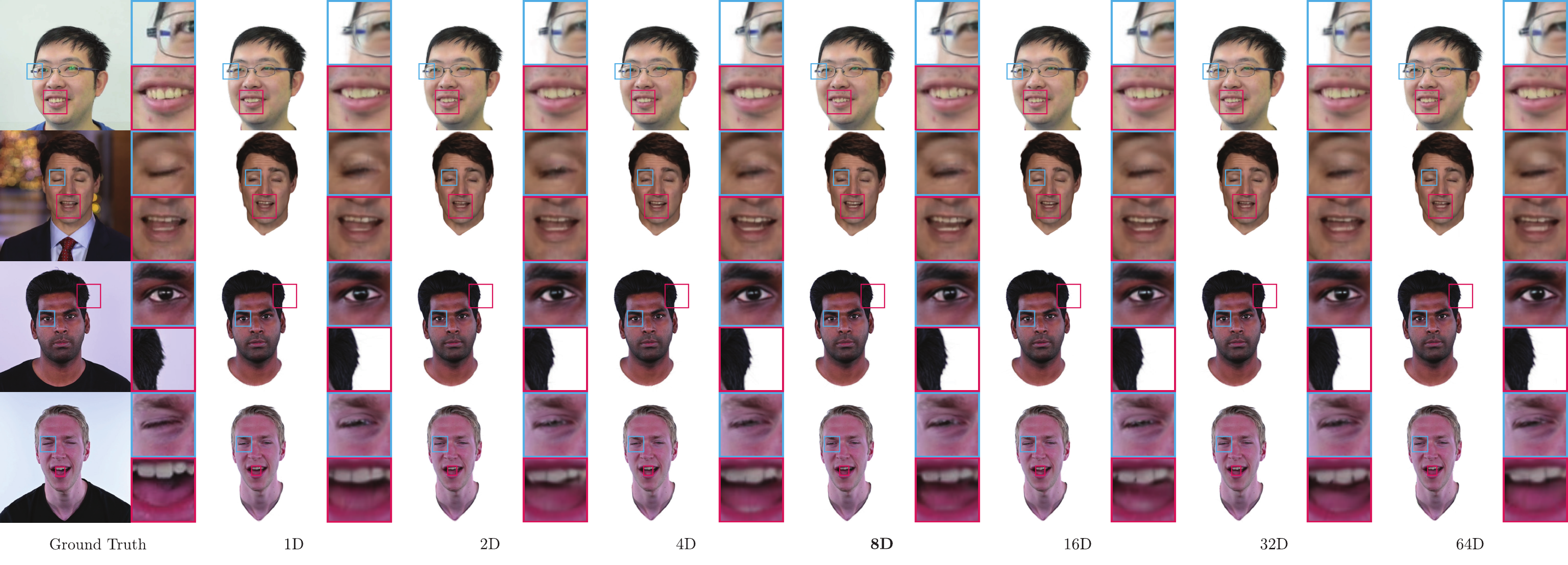}
  \caption{\textbf{Qualitative comparison} for varying latent dimensionalities. We find that \oursname{}s are robust towards different latent dimensions. A latent dimension of $8$ performs best, but we already observe an improvement for a single latent dimension ($n=1$) over the vanilla 3DGS variant.
  \label{suppfig:ablation}}
\end{figure*}

\begin{figure*}[ht]
\begin{center}
\small
\setlength{\tabcolsep}{2pt}
\newcommand{\inputwidth}{3cm}
\begin{tabular}{rccccc}
\rotatebox{90}{FlashAvatar} &
  \includegraphics[width=\inputwidth]{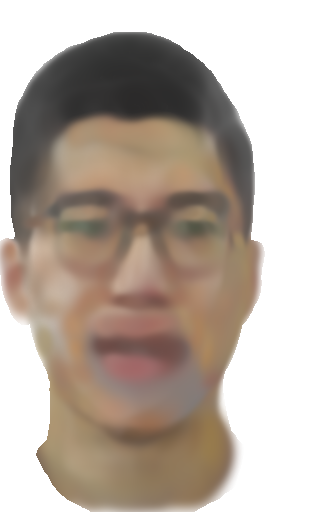}  &
      \includegraphics[width=\inputwidth]{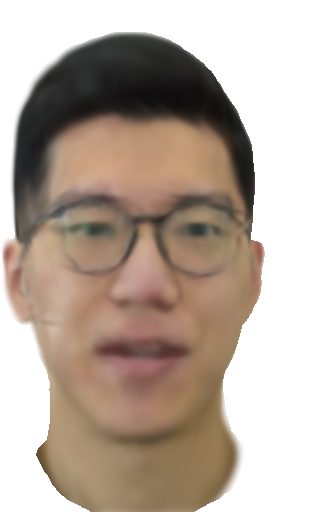}  &
      \includegraphics[width=\inputwidth]{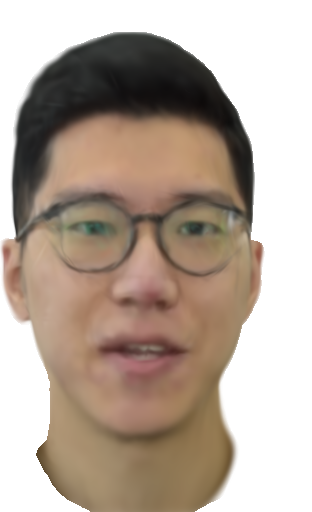}  &
      \includegraphics[width=\inputwidth]{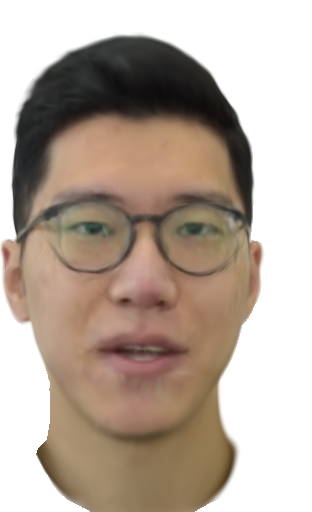}  &
      \includegraphics[width=\inputwidth]{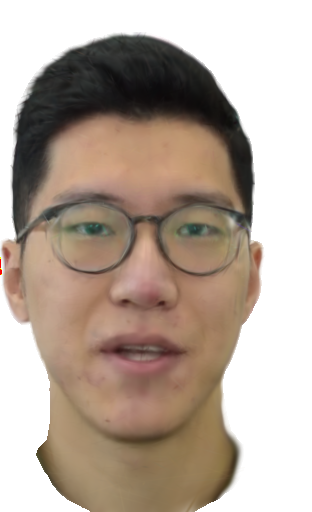} \\
\rotatebox{90}{\textbf{Ours}} &
  \includegraphics[width=\inputwidth]{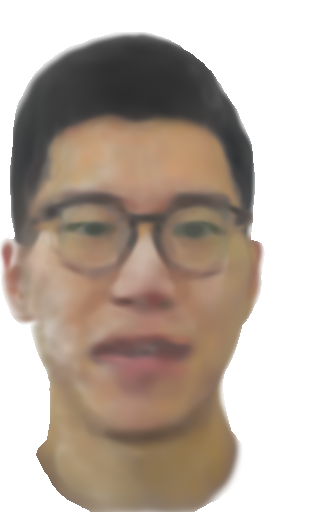}  &
      \includegraphics[width=\inputwidth]{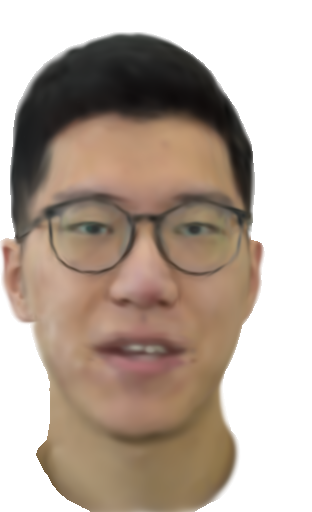}  &
      \includegraphics[width=\inputwidth]{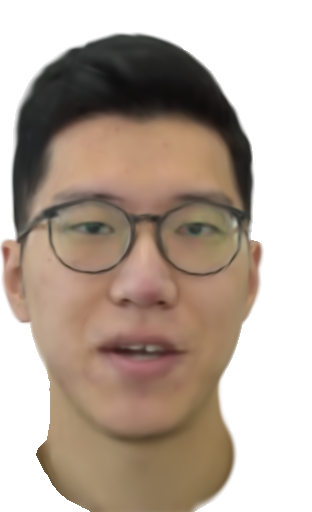}  &
      \includegraphics[width=\inputwidth]{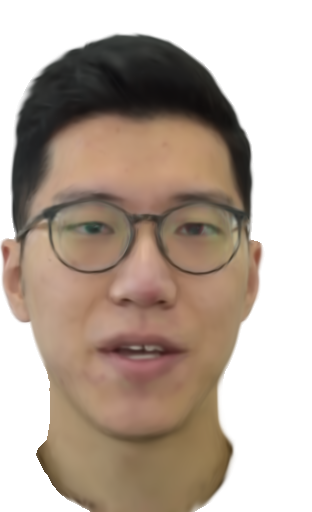}  &
      \includegraphics[width=\inputwidth]{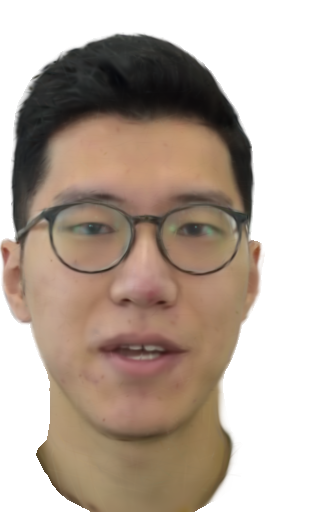} \\
\rotatebox{90}{FlashAvatar} &
  \includegraphics[width=\inputwidth]{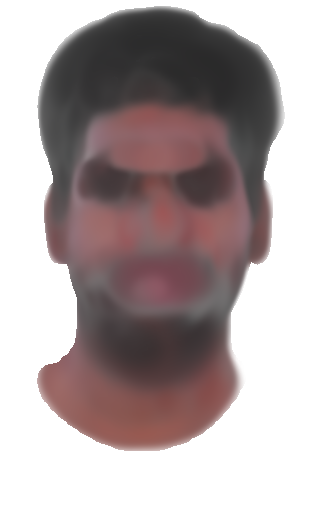}  &
      \includegraphics[width=\inputwidth]{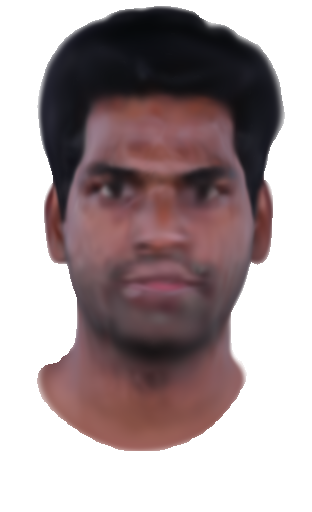}  &
      \includegraphics[width=\inputwidth]{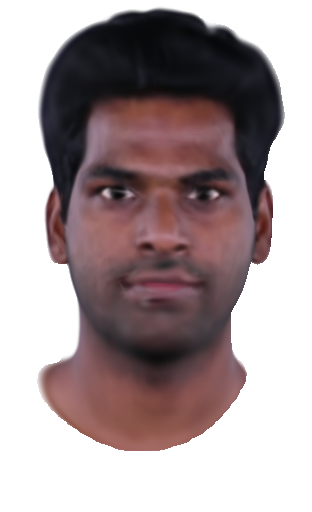}  &
      \includegraphics[width=\inputwidth]{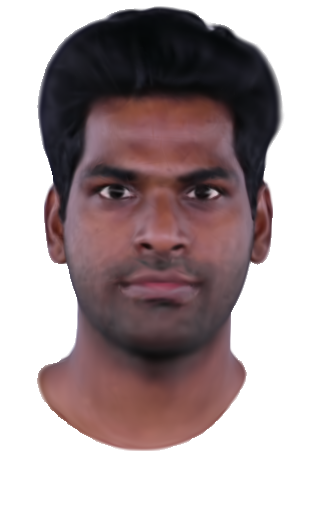}  &
      \includegraphics[width=\inputwidth]{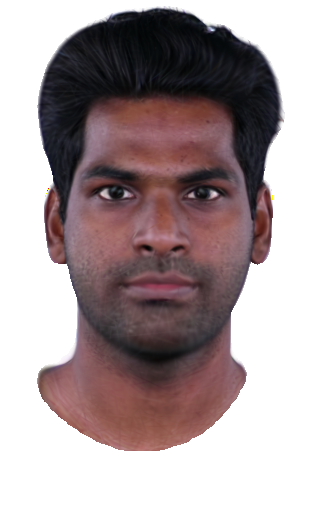} \\
\rotatebox{90}{\textbf{Ours}} &
  \includegraphics[width=\inputwidth]{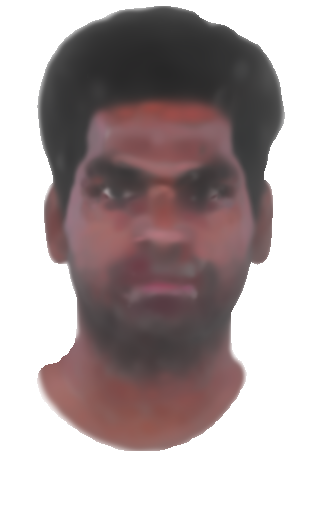}  &
      \includegraphics[width=\inputwidth]{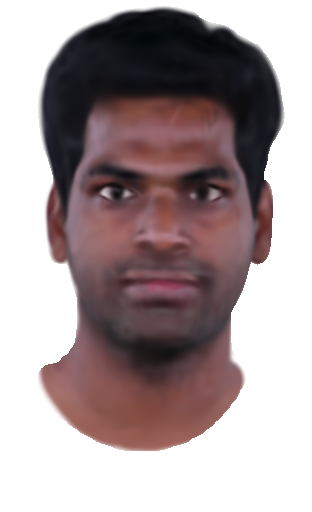}  &
      \includegraphics[width=\inputwidth]{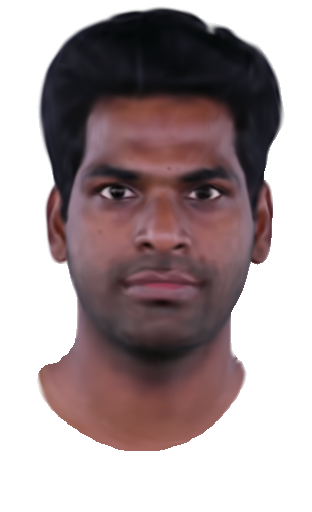}  &
      \includegraphics[width=\inputwidth]{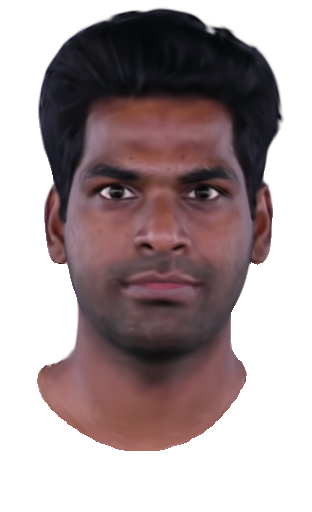}  &
      \includegraphics[width=\inputwidth]{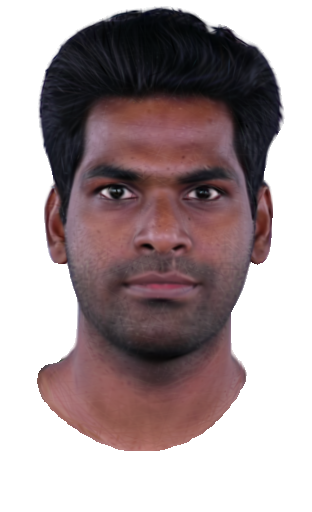} \\
& 5 seconds & 20 seconds & 60 seconds & 180 seconds & 600 seconds \\
\end{tabular}
\end{center}
\caption{\label{suppfig:training_time_vs_result}We compare the convergence speed of FlashAvatar \cite{xiang2024flashavatar} vs. \textbf{Ours}. The \emph{only difference} between FlashAvatar and \textbf{Ours} is the substitution of 3D Gaussians (top) with \oursname{}s (bottom), as described in the case study in the main paper. From the beginning, \oursname{}s display sharper results.
}
\end{figure*}

We complement the ablation study from the main paper with supplementary results for different MLP configurations in \cref{supptab:quantitative_comparison_ablation} and \cref{suppfig:ablation}. The default FlashAvatar MLP \cite{xiang2024flashavatar} has 6 layers with 256 neurons, totaling 375K parameters. Replacing vanilla 3D Gaussians with \oursname{}s adds optimizable parameters (see \cref{suppssec:parameteriztion}).
One might assume that simply increasing the parameter count for the FlashAvatar MLP would improve the results. However, this is not the case. 
We ablate different MLP configurations in \cref{supptab:quantitative_comparison_ablation}. Adding more parameters to the MLP does not perform as well as adding \oursname{}. In fact, it performs the same, or worse than the baseline, while significantly slowing down rendering speed. FlashAvatar with vanilla 3DGS runs at $347$ FPS. With a large MLP, this number drops to $158$ (for $256\times 40$, 2.6M parameters) and $178$ (for $512 \times 11$, 2.7M parameters). With \oursname{}s ($n=8$ and 2.6M parameters), the original MLP ($256 \times 6$) outperforms the other MLP variants for all metrics while maintaining a rendering speed of $300$ FPS. All metrics and rendering times were computed on a single NVIDIA GeForce RTX 2080 Ti for images with resolution $512 \times 512$. In summary, the \oursname{}s' performance improvement cannot be matched by increasing the complexity of the MLP. \oursname{}s boost the performance while maintaining fast rendering speed. Moreover, while FlashAvatar with vanilla 3DGS achieves higher FPS, we observe faster convergence with \oursname{}s. \cref{suppfig:training_time_vs_result} visualizes the training progress over time and shows that \oursname{}s achieve a higher quality for a given time budget.

A fundamental advantage of \oursname{}s is their ability to distill highly local context, enabling independent deformations between spatially proximate but semantically distinct regions. For instance, our method can independently model glass frames near the upper cheek or the upper teeth adjacent to the jaw. In contrast, FlashAvatar suffers from stronger coupling between neighboring Gaussians due to its shared MLP architecture and direct offset approach. This coupling creates an optimization challenge where improvements in one region often degrade quality in others. Our approach allows each region to optimize independently, preserving detailed geometry and appearance across semantically different but spatially adjacent facial features.

\label{supssec:ablation_study}
\begin{table}
    \centering
    \resizebox{\columnwidth}{!}{\begin{tabular}{cccc|ccc}
        MLP (width$\times$depth) & HGS-Dim. & \# Param. & FPS & PSNR ↑ & SSIM $\bigl(10^{-1}\bigr)$ ↑ & LPIPS$\bigl(10^{-2}\bigr)$ ↓ \\
        \hline
        256$\times$6 & - & 375K & 347 & 29.43 & 9.466 & 5.107 \\
        256$\times$40 & - & 2.6M & 158 & 28.10 & 9.380 & 5.720  \\
        512$\times$11 & - & 2.7M & 178 & 29.50 & 9.472 & 5.122\\
        \hline
        256$\times$6 & 1D & 1.2M & 291 & 29.73 & 9.492 & 5.066 \\
        256$\times$6 & 2D & 1.4M & 304 & \cellcolor[HTML]{FFF2CC}29.92 & 9.503 & 5.000 \\
        256$\times$6 & 4D & 1.8M & 293 & 29.89 & 9.507 & 4.994  \\
        256$\times$6 & 8D & 2.6M & 300 & \cellcolor[HTML]{E2EFDA}29.99 & 9.510 & \cellcolor[HTML]{FFF2CC} 4.978 \\
        256$\times$6 & 16D & 4.1M & 298 & \cellcolor[HTML]{FFF2CC}29.92 & 9.511 & \cellcolor[HTML]{FFF2CC} 4.978  \\
        256$\times$6 & 32D & 7.2M & 281 & 29.89 & \cellcolor[HTML]{FFF2CC} 9.511 & \cellcolor[HTML]{E2EFDA}4.976 \\
        256$\times$6 & 64D & 13.4M & 273 & 29.91 & \cellcolor[HTML]{E2EFDA}9.512 & 5.020 \\
    \end{tabular}}
    \caption{\textbf{MLP and Latent Dimensionality Ablations.} Simply increasing the parameter count for the FlashAvatar MLP does not improve the metrics. Our \oursname{}s, however, improve the performance of the original MLP out-of-the-box. As an additional benefit, \oursname{}s render at around $300$ FPS while the deeper MLPs, with comparable parameter counts, drop to $158$ FPS ($256 \times 40$) and $178$ FPS ($512 \times 11$), respectively. Note that the drop of $10$ FPS for 1D is likely due to memory bottlenecks caused by poor cache and vector load/store locality.
    \colorbox[HTML]{E2EFDA}{Green} denotes the best and \colorbox[HTML]{FFF2CC}{Yellow} the second best.}
    \label{supptab:quantitative_comparison_ablation}
\end{table}

We ablate the effect of different latent dimensionalities in \cref{supptab:quantitative_comparison_ablation}. We find that \oursname{} are robust towards different latent dimensions. A latent dimension of $8$ performs very well, but we already observe an improvement for a single latent dimension ($n=1$) over the vanilla 3DGS variant. The top row corresponds to the FlashAvatar baseline \cite{xiang2024flashavatar}, which does not use any \oursname{}s.

\section{Societal Impact}
\label{suppsec:impact}
It is important to be aware that photorealistic, high-quality face avatars from monocular videos can have societal implications. While our novel \oursname{} representation offers exciting possibilities for entertainment, communication, and virtual experiences, it could potentially be misused to spread misinformation and deception. Realistic face avatars could be exploited to produce convincing deepfakes, potentially undermining trust in visual media and influencing societies and politics. We strongly condemn any form of abuse or malicious use of our research and advocate for responsible development and application of face avatar technology, always in strict accordance with local laws and regulations.

\end{document}